\documentclass{article}
\usepackage{iclr2027_conference,times}
\usepackage{wrapfig}
\usepackage{graphicx}
\usepackage{amsmath}
\usepackage{amssymb}
\usepackage{booktabs}
\usepackage{placeins}
\usepackage{url}
\usepackage{hyperref}
\usepackage{longtable}
\usepackage{float}

\title{Signatures of semantic search in the activations of large language models}

\author{
  Luke Leckie$^{1,2,3}$ \quad
  Peter M. Todd$^{2,4}$ \quad
  Jacob G. Foster$^{1,2,3,5}$ \\[0.5em]
  $^{1}$Department of Informatics, \\
  Luddy School of Informatics, Computing, and Engineering, \\
  Indiana University, Bloomington, IN \\
  $^{2}$Cognitive Science Program,
  Indiana University, Bloomington, IN \\
  $^{3}$Center for Possible Minds,
  Indiana University, Bloomington, IN \\
  $^{4}$Department of Psychological and Brain Sciences, \\
  Indiana University, Bloomington, IN \\
  $^{5}$Santa Fe Institute, Santa Fe, NM \\[0.5em]
  Correspondence: \texttt{lleckie@iu.edu}
}
\iclrfinalcopy
\begin{document}
    \maketitle
    \fancyhead{}
    \begin{abstract}

When recalling lists of concepts (e.g., animals) during the semantic fluency task (SFT), both humans and large language models (LLMs) organise their output into clusters of related items (e.g., sea animals) that are punctuated by strategic switches between clusters. In humans, this pattern can be explained by a semantic foraging process, whereby distinct neural and behavioural signatures accompany within-cluster production (“exploit") and between-cluster switching (“explore"). Whether LLMs likewise represent these two search regimes within their internal states is unknown. Here, we apply a range of mechanistic interpretability techniques to provide evidence for this. In Study 1, we use the Jacobian lens (J-lens), which maps intermediate-layer residual-stream representations to token-level activations, to show that concept-level activations predict switching. First, we find that switching coincides with low next-token activations. Moreover, the probability of switching rises as the set of strongest J-lens activations (the J-space) becomes depleted of items from the category currently being produced, analogous to explore-exploit decision-making during patch foraging. We then show that middle-layer J-lens activations of abstract category-related labels (e.g., “water") increase in anticipation of switching into that category. We confirm these representations to causally influence switching by deriving steering vectors that target category switching. In Study 2, we identify generic residual stream directions that are activated during and in anticipation of switching. By steering activations along these directions, we bias increased or decreased rates of switching. Our study extends the semantic foraging framework to artificial intelligences and provides evidence that LLMs maintain distinct representational signatures for exploration and exploitation as they verbalise conceptual information.

    \end{abstract}
    \vspace{1em}

\section{Introduction and related work}
Searching for resources is a ubiquitous challenge faced by agents \citep{hillsExplorationExploitationSpace2015a, charnov1976optimal, todd2012cognitive}. In search environments where resources are distributed patchily over space, agents must balance the local exploitation of known but depleting resources against global exploration for novel but uncertain resources \citep{charnov1976optimal}. This type of search problem is typified by animal patch foraging behaviours. During patch foraging, animals typically decide when to leave a depleting patch and explore for a new one based on the value of continued exploitation relative to expected opportunities available elsewhere \citep{charnov1976optimal}. Alternation between local exploitation and global exploration has been described across a wide diversity of organisms and contexts; it may constitute a general solution to efficient search in structured environments \citep{marques2020internal,nicolau2009directional,hogeveenNeurocomputationalBasesExploreexploit2022,todd2012cognitive,buatois2016evidence}.

Mounting evidence from human concept-list recollection studies suggests that the retrieval of semantic information from internal meaning spaces also operates according to a foraging process that balances local exploitation against global exploration \citep{hills2012optimal,lundin2023neural,nour2023trajectories}. Research has focused on the Semantic Fluency Task (SFT), in which individuals are tasked with naming as many concepts (e.g., animals) as they can within a short period of time.  Concept recollection during the SFT is typically arranged into bouts of exploitative “clustering", in which individuals name closely semantically related items (e.g., listing farm animals); these bouts are separated by strategic exploratory “switches" between clusters (e.g., farm animals $\rightarrow$ sea animals) \citep{troyer1997clustering,lundin2020semantic}. The semantic foraging framework is supported by studies showing that clustering and switching regimes are coupled to distinct behavioural and neural signatures. For instance, when switching, inter-item response latency increases \citep{hills2012optimal} and both the posterior cerebellum and hippocampus experience heightened activation, with characteristic “ripple bursts" of activity travelling through the hippocampus \citep{lundin2023neural,nour2023trajectories}. Moreover, the activity of these two regions progressively increases (“ramping") with the time spent naming items within a given cluster \citep{lundin2023neural}. Finally, evidence for patch-like semantic representations comes from work showing that priming individuals with high-level category-related concepts (e.g., “river”) facilitates the retrieval of exemplar items (e.g., “crocodile”), suggesting that item-level retrieval depends on the activation of category-level semantic representations \citep{neelySemanticPrimingRetrieval1977,diraniTimeCourseCrossmodal2023,rissmanWordsNotJust2024}. While SFT-performing Large Language Models (LLMs) similarly produce concept lists arranged into clusters and switches \citep{lacosseEmergingHumanlikeStrategies2026a} it is not presently known if this is related to a semantic foraging-like process---with distinct activation signatures for clustering and switching---or only bears superficial resemblances to human-produced text. 

Recent mechanistic interpretability research provides evidence that LLMs maintain internal representations that track and influence the structure of generated language \citep{palFutureLensAnticipating2023,hendelInContextLearningCreates2023,toddFunctionVectorsLarge2024a,dongEmergentResponsePlanning2025}. For example, research using sparse autoencoders to track changes in model activations when processing pre-created stories finds that activations strongly shift across event boundaries \citep{lubanaPriorsTimeMissing2025}. Furthermore, research examining LLM reasoning behaviours, such as back-tracking and expressing uncertainty, has found that these behaviours are coupled to specific directions in model activation space, which can be “steered" to bias the propensity for these behaviours \citep{venhoffUnderstandingReasoningThinking2025a}. Other research provides evidence that models hold and act on anticipatory internal representations of future content. Specifically, LLMs writing rhyming poetry internally represent candidate end-of-line rhyming words before composing the line; editing these representations restructures the intervening text to cohere with the edited end-of-line word\citep{lindsey2025biology}.  More recently, work employing the “Jacobian-lens" (J-lens), which maps intermediate-layer activations to token-level activations, has found that LLMs can represent intermediate and higher-level concepts related to their current output in their mid-layers (e.g., the token “red" may have a high activation when the LLM is answering “The fourth planet from the sun is [Mars]") \citep{gurnee2026verbalizablerepresentationsformglobal}. One recent study found that LLMs processing human-generated SFT lists assign low next-token probabilities to their switch events; these LLMs also have different residual stream activations when specifically prompted to generate a semantically similar versus different item to one provided \citep{lacosseEmergingHumanlikeStrategies2026a} and can collaborate with humans performing the SFT to enhance their concept retrieval \citep{lacosse2026artificial}. Together, these findings link changes in generated language to identifiable internal representations. However, research has not yet directed mechanistic interpretability tools, such as the J-lens, to study or manipulate the dynamics of LLM activations as they naturally perform semantic foraging tasks. It is, therefore, unclear if the representations of LLMs support a foraging-like process that regulates transitions between exploitation and exploration during semantic retrieval. Understanding this is critical:  The same area-restricted search dynamics that govern SFT production are hypothesised to extend into naturalistic speech \citep{leckiePsychosisassociatedDisruptionsSemantic2026} and could also be shared across a range of search tasks that are central to the utility of LLMs, including searching for information over the internet \citep{nakanoWebGPTBrowserassistedQuestionanswering2022}, selecting tools \citep{schickToolformerLanguageModels2023}, navigating code \citep{yangSWEagentAgentComputerInterfaces2024}, and retrieving information from context \citep{liuLostMiddleHow2024}. Discovering latent explore-exploit dynamics in these behaviours and modulating the balance of exploration versus exploitation could enhance the efficiency of the search processes underpinning the effectiveness of these LLM behaviours.

In this paper, we apply a range of mechanistic interpretability techniques to study this. In Study 1, we examine how concepts are represented across different layers of LLMs during the SFT and ask if these representations signify participation in switching versus clustering behaviours. To do this, we first investigate how final-layer token probabilities differ preceding switching versus clustering events. We then test whether final-layer differences in concept activation are detectable in earlier layers, by applying the J-lens to internal layers of the residual stream. We find that switch events are preceded by low next-token probabilities and that, although all items show increased activation in anticipation of their verbalisation, this effect is lower for items participating in switches, as compared to clustering. Second, we use the J-lens to identify internal conceptual patches comprising the most strongly activated concepts. We find that as these patches become depleted with items from the same category that an LLM is presently verbalising from, LLMs become increasingly likely to switch to a novel category; this effect is greater than expected by depleting the overall pool of items in a category.  Third, we show that the activations of higher-level category-related labels (e.g., “Africa") increase in anticipation of switching into the corresponding category. We then demonstrate these item- and category-level activations to influence switching behaviour by using them to derive steering vectors that induce switching towards target categories. In Study 2, we identify generic directions in LLM activation space that predict switching behaviour and ramp in anticipation of switching. By steering SFT-performing LLMs along this direction, we bias increased or decreased likelihoods of switching. This study provides evidence that LLMs hold distinct representational signatures for exploration and exploitation as they search for conceptual resources, demonstrating semantic foraging-like search behaviour as a property of LLMs.

\section{Experimental set-up}
Our experiments were replicated across five open-weight instruction-tuned LLMs, ranging between 2-billion and 9-billion parameters: Gemma-2-9B (42 layers), Gemma-2-2B (25 layers), Llama-3.1-8B (32 layers), Llama-3.2-3B (28 layers), and Qwen2.5-7B (28 layers). We note that even the smallest models are effective at performing the SFT, given the simplicity of the task. For each model, we performed experiments analysing or steering model activations at regularly spaced residual stream layers (every second layer, from layer 2 to near-final depth: layers 2–40 for Gemma-2-9B, 2–24 for Gemma-2-2B, 2–26 for Qwen2.5-7B,  2–30 for Llama-3.1-8B, and 2–26 for Llama-3.2-3B). To ease model comparison, in our figures we report fractionalised layer depths, with 0 being the initial input layer and 1 being the final readout layer. For each model-layer combination, experiments were iterated over 100 different random seeds. All generations used nucleus sampling with temperature=0.9 and top-p=0.95, with a fixed random seed. All prompts were issued via each model's own chat template, without a system prompt. Outputs were capped at 200 tokens.

All experiments tasked LLMs with performing the Semantic Fluency Task (SFT) (Fig. 1A). To do this, LLMs were prompted as follows:
\textit{"Name as many different animals as you can, one after another, separated by commas. Just the list."}

We designated animal categories by using an extended version of the well-established Troyer category designations \citep{troyer1997clustering}. In these category designations, animals can participate in multiple categories (e.g., alligator belongs to both reptiles/amphibians and water). Switching is identified when a concept is named that shares none of the category labels with the preceding concept (see Appendix for further details). 
\begin{figure*}[t]
  \centering
  \includegraphics[width=1\textwidth]{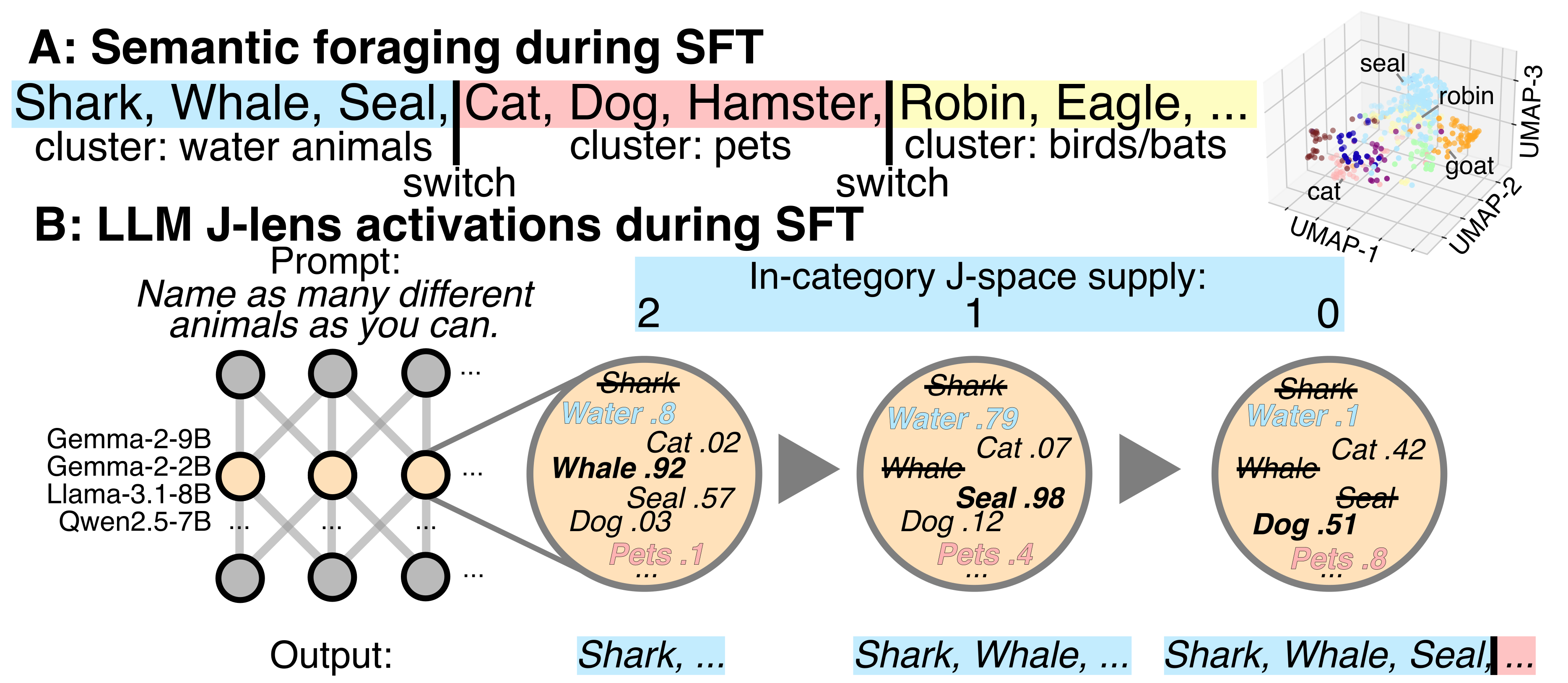}
  \caption{\textbf{Semantic fluency and LLM J-lens activations.} \textbf{A.} Left: Example Semantic Fluency Task (SFT) output. Cluster labels are given by colours and switches between clusters are indicated by vertical black lines in-between distinct clusters. Right: Uniform Manifold Approximation (UMAP) of Word2Vec embeddings of animal names. Colours indicate extended Troyer category designations. \textbf{B.} Illustration of LLM J-lens activations during SFT. Left neural network represents an LLM. Beige-highlighted layer represents residual stream representations of a layer under study by the J-lens. Beige circle containing concepts represents a set of concepts included within the J-space for that layer. Note that in our experiments we considered the J-space as the top-25 strongest J-lens activations. J-lens activations are indicated by numbers to the right of concepts. Items that have already been verbalised are indicated by strikethroughs. Higher-level abstract category-related labels (“water" and “pet") are indicated in blue and red font, respectively. The in-category J-space supply is computed as the number of items (not including category-related labels) within the J-space that belong to the category from which the model is presently naming items but that have not been verbalised yet. In this example, as the LLM names water animals and depletes the in-category J-space supply for that category, it switches to naming animals within the pet category. Note that the J-lens activation for the abstract category-related “pet" label increases in anticipation of switching to that category (See Figure 3 for analysis).}
  \label{fig:experiment_illustration}
\end{figure*}

\section{Study 1: Concept activations indicate and anticipate switching}
\subsection{Item activations differ between clustering and switching }
In human semantic fluency, switching behaviour is thought to be tied to the exploration for new items and semantic categories \citep{hills2012optimal}. In Study 1, we aim to test for a similar exploratory process in SFT-performing LLMs by comparing the activations of individual concepts during clustering and switching. To do this, we first ask if the next-token probabilities for verbalised items, as computed at the final layer, differ at the token position immediately prior to a switching event, as compared to a clustering event (e.g., “shark" may have a different next-token probability in the context of a switch versus a cluster). Semantic foraging expects that next-token probabilities should be lower in advance of switching, reflecting an exploratory search over next items. Consistent with this prediction and work examining this in human-produced SFTs \citep{lacosseEmergingHumanlikeStrategies2026a}, next-token probabilites were significantly lower for switch-participating items, as compared to cluster-participating items (Fig. 2A, Linear Regression [LR], all models: $\chi^2\geq 32.39$, $p<0.0001$). 

\begin{figure*}[t]
  \centering
  \includegraphics[width=1\textwidth]{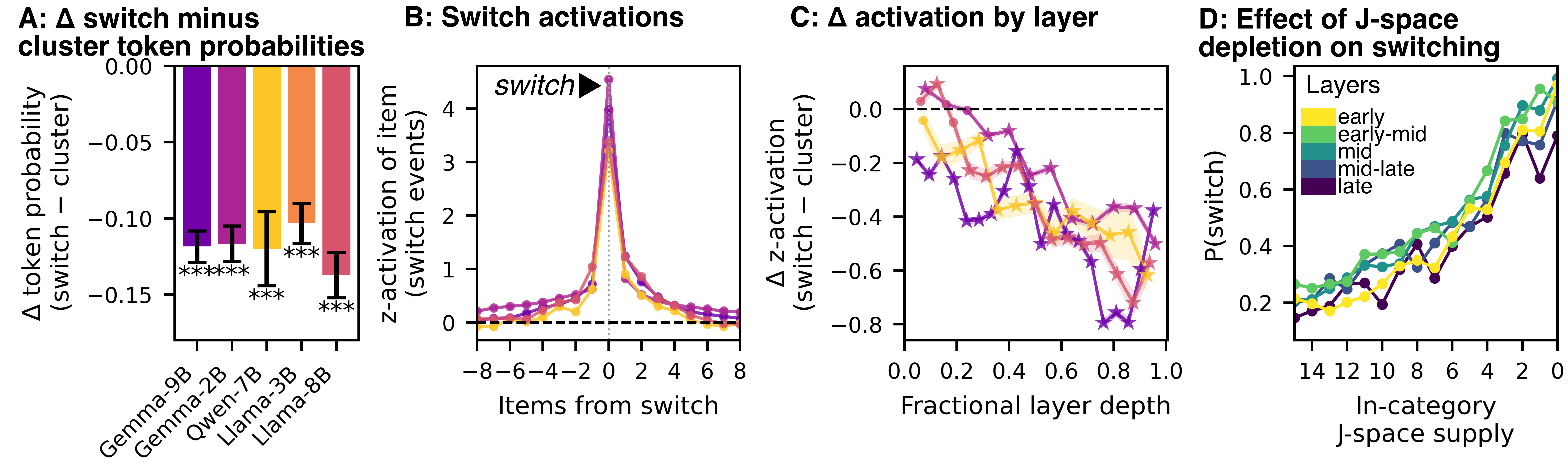}
   \label{fig:surprisal_jlens}
  \caption{\textbf{Semantic fluency item activations and J-space depletion.} \textbf{A.} Comparison of next-token probabilities for items participating in clustering versus switching events across models. Coloured bars extend to the average difference in next token probabilities for switching events minus clustering events. Black whiskers extend to $\pm$standard error (se). Next-token probabilities were significantly lower during switching (linear models, all p$<$0.0001***) for all models. \textbf{B.} Z-scored J-lens activations for items participating in switching, measured in advance (negative lag) and after (positive lag) their emission (lag=0). Points represent means across seeds and layers; shaded areas represent standard errors. \textbf{C.}  $\Delta$ in Z-scored J-lens activations across fractionalised layers, as measured immediately prior to item emission (lag=-1). Points represent means across seeds and bars around points extend to $\pm$se. Stars for means indicate significantly different activations (p$<$0.05 after Benjamini-Hochberg [BH] correction) of item activations between items participating in switching versus clustering, as tested by linear models. \textbf{D.} The probability of switching to a new category as a function of the in-category J-space supply, averaged across layer blocks and models. More yellow points indicate earlier layer blocks and purple, later. Note that J-space supply \textit{decreases} from left to right.}
 
\end{figure*}
We then ask if switch-associated differences in final-layer model activations are detectable in earlier residual stream layers. To derive item-level activations from residual stream layers, we use model-specific Jacobian-lenses (J-lens). As described in other work \citep{gurnee2026verbalizablerepresentationsformglobal}, the J-lens allows the inference of item-level activations from intermediate residual stream layers by mapping intermediate representations to final-layer states. The J-lens activations of a concept roughly correspond to the tendency of a model to verbalise that concept, or items related to that concept (e.g., “sea" could be highly activated as the model prepares to say “dolphin"). We use pre-computed J-lenses, available on Neuronpedia \citep{neuronpedia}. As Llama-3B does not have a precomputed J-lens, we did not perform J-lens experiments with this model.

Consistent with our token probability findings (Fig. 2A), we expect that the activations of switch-participating items should be lower than that of cluster-participating items. Indeed, we find that although the activations of both cluster and switch participating items increase in anticipation of their emission (Fig. 2B; Fig. S1), this effect is weaker for switching than for clustering (Fig. S1, Table \ref{tab:J-lens-activations-lag}) and activations are significantly lower for switch-participating items, as compared to cluster-participating items prior to their emission (Fig. 2C, Table \ref{tab:jlens}, LR, all models from layer 8 up: $\chi^2\geq 98.17$, $p<0.0001$). For most models this effect is apparent from the early-middle layers and increases in potency up until the late layers (Fig. 2C). This result provides evidence that elevated next-item uncertainty, characterising semantic exploration, is represented throughout LLM layers.

\subsection{J-space concept depletion predicts switching}
The semantic foraging framework predicts that as a given category, or patch, of semantic information becomes depleted, a strategic decision to switch to a new category (“explore") should become increasingly likely \citep{hills2012optimal,charnov1976optimal}. Therefore, we next ask if the J-lens can be used to identify conceptual patches within the activations of LLMs and whether their depletion predicts switching behaviour (Fig. 1). Previous work has defined the concepts corresponding to the top-25 strongest J-lens activations as comprising a “Jacobian-space" (J-space), which has been compared to the global workspace model of human cognition \citep{gurnee2026verbalizablerepresentationsformglobal}. In our experiments, we therefore test if the number of items in the J-space that belong to the current category and have not been verbalised predicts switching (Fig. 1B). If LLMs perform semantic foraging, then as the in-category J-space supply drops, the probability of switching to a new category should increase.

As expected by the semantic foraging framework, we find that the probability of switching increases sharply as the J-space supply drops (Fig. 2D, Table \ref{tab:j-spacedepletion_base}, LR, for all models in middle layer blocks: $r\leq 0.31$, $p<0.0001$). Furthermore, the effect of J-space depletion on switching probability is significantly stronger than when considering the effect of depleting the overall pool of animals that could be named within a given category for all models except for Qwen (Table \ref{tab:j-space_supply}, LR, effect at layer block with strongest effect: $F\geq 19.94$, $p<0.0001$). These effects are robust across tested J-space sizes (See Appendix, Table \ref{tab:j-space_robustness}) and are generally most salient in the middle layers, consistent with previous work interpreting the J-space as emerging within this layer range \citep{gurnee2026verbalizablerepresentationsformglobal}.

\subsection{Activation of abstract category-related labels anticipates switching }
During human semantic fluency, the retrieval of individual items that are members of a given category is thought to depend upon the activation of that category's semantic representation (e.g., naming “shark" and “jellyfish" depends upon the (latent) activation of the “sea animal" category) \citep{neelySemanticPrimingRetrieval1977,diraniTimeCourseCrossmodal2023,rissmanWordsNotJust2024}. Thus, we next ask if LLMs similarly demonstrate heightened activation of abstracted category-related labels while clustering within that category---or in anticipation of switching into that category. To do this, we identify a number of category-related labels (Fig. 3E) and: 1) compare their average J-lens activations during clustering within that category, as compared to clustering in an unrelated category (Fig. 3A); 2) measure their J-lens activations before and after switching into a target/non-target category (Fig. 3B-D). 

Consistent with the idea that category activation is coupled to the verbalisation of exemplar items in LLMs, we find that J-lens activation for category-related labels is heightened when models cluster within that target category as compared to unrelated categories; this effect is strongest in the middle layers (Fig. 3A, Table \ref{tab:category_stats}). The activation of labels progressively increases in anticipation of switching into their target category and spikes at the switch event, before progressively decreasing  (Fig. 3B, LR, for all models, effect at switch into target category : $F\geq 848.35$, $p<0.0001$). Interestingly, we observe that the activations of labels unrelated to the category being switched to also tend to increase in anticipation of switching (although this effect is weaker than for target labels \ref{tab:category_stats}), but that their activations drop sharply \textit{at} the switch event before increasing again (Fig. 3C-D). This suggests that category selection during switching could be driven by competition between the activations of various categories. One category eventually receives higher relative activation and becomes the “winner" that gets switched to; the “losing" categories get suppressed (Fig. 3B-C-D).

\begin{figure*}[t]
  \centering
  \includegraphics[width=1\textwidth]{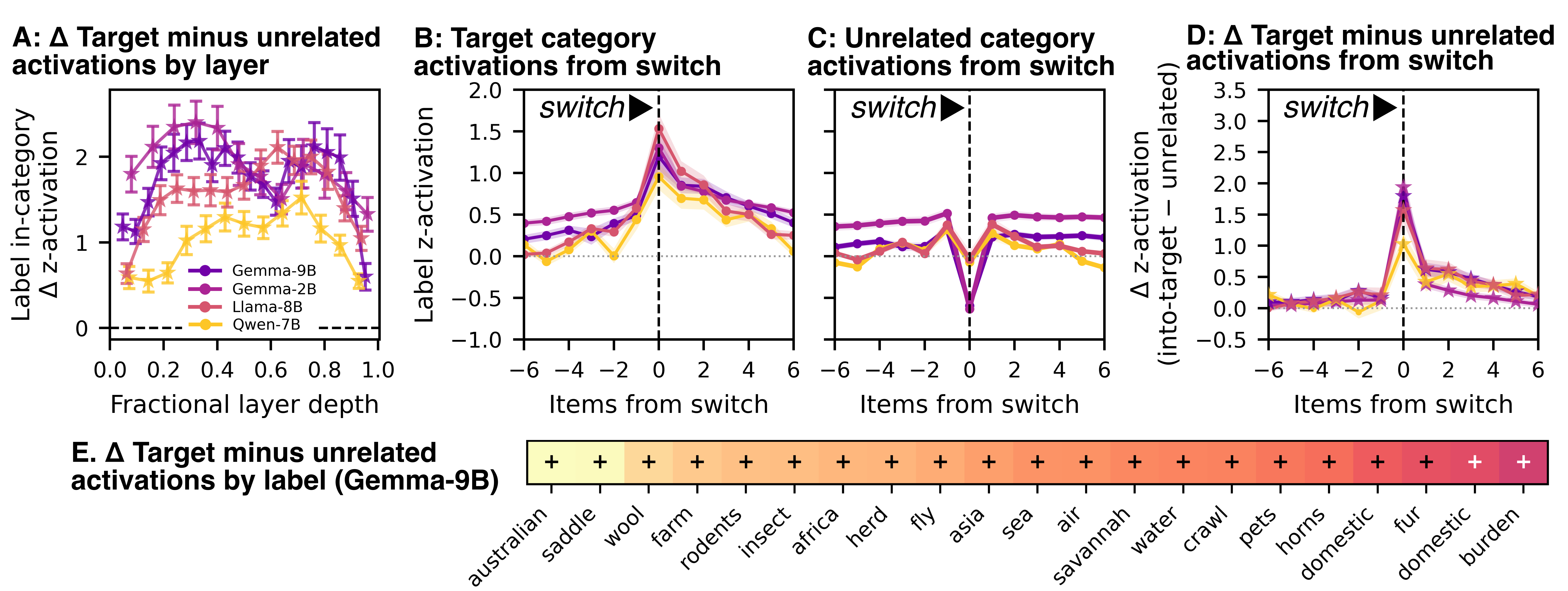}
  \caption{\textbf{J-lens activations of category-related labels during LLM semantic fluency.} \textbf{A.} $\Delta$ in the J-lens activations of category labels, subtracting the mean activation of each label during production of items in unrelated categories from mean activation during production of items in clusters with that category label (target category). Points indicate means across all tested labels and seeds. Bars extend to $\pm$se. \textbf{B-D.} Layer-averaged J-lens activations of category labels preceding (negative x-values) and following a switch (positive x-values) into the target category (\textbf{B}) or an unrelated category (\textbf{C}), as well as the $\Delta$ between the two (target minus unrelated activation) (\textbf{D}). Points indicate means across seeds and layers; shaded areas indicate standard errors. \textbf{A} and \textbf{D}: Stars for means indicate significantly different activations (p$<$0.05 after BH correction) of category-related labels between target and unrelated categories. \textbf{E.} $\Delta$ in target minus unrelated category label activations for each tested label (x-axis) for Gemma-2-9B, averaged across layers and seeds. More positive values (lighter cells) indicate stronger $\Delta$ activation of that label. Note that the first instance of “domestic" maps to the “domestic animals" category and the second maps to the “pets" category.  }
  \label{fig:category_anticipation}
\end{figure*}

\subsection{Concept activations have causal influence on switching}
\begin{figure*}[t]
  \centering
  \includegraphics[width=1\linewidth]{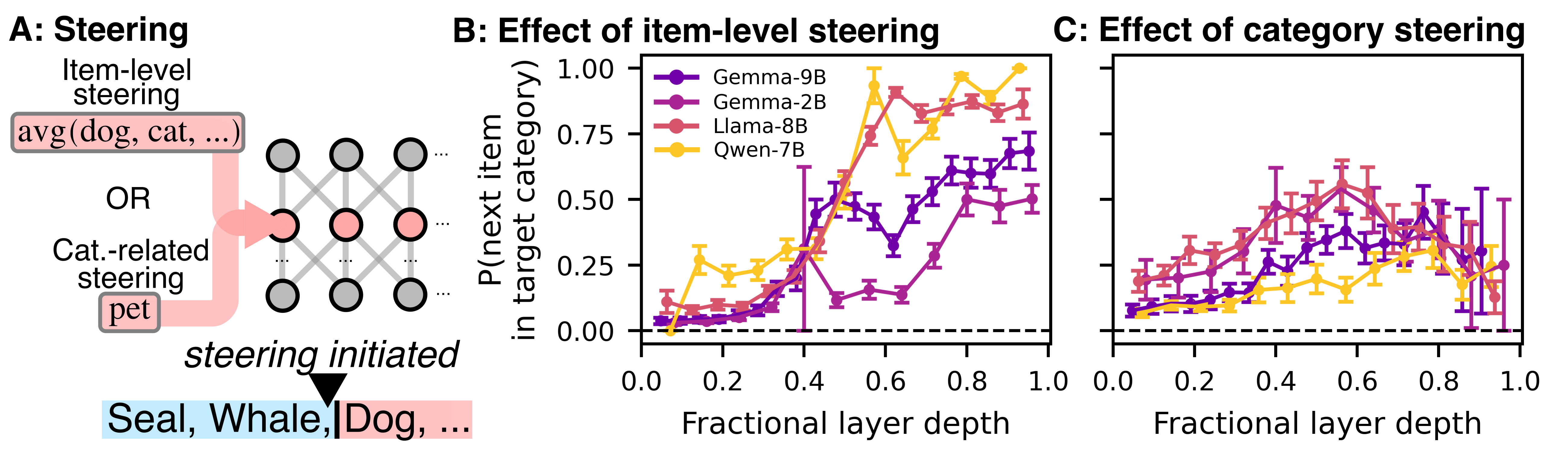}
  \caption{\textbf{Steering model activations to induce switching into target categories.} \textbf{A.} Models were induced to switch into target categories by steering vectors generated from averaging activations over item-level representations or from individual abstracted category-related labels. \textbf{B-C.} Effect of steering using item (\textbf{B}) or category-related (\textbf{C}) steering vectors. Points represent means across target-level probabilities computed across seeds and target categories tested and bars extend to $\pm$se. }
  \label{fig:cat_steering}
\end{figure*}

So far, our study provides evidence that the activations of LLMs show distinct signatures preceding switching events. This is true not only of the individual items that will be verbalised as part of the SFT (Fig. 2) but also for their higher-level category-related labels (Fig. 3). We next test if these signatures correspond to causal influences on switching behaviour during SFT. To do this, we create two different types of steering vectors (Fig. \ref{fig:cat_steering}A): 1) \textit{Item-level} steering vectors that average the residual stream activations of all individual items that fall within a target category; and 2) \textit{Category-related} steering vectors that are the residual stream activations of one of the higher-order category-related labels used in the previous analysis.
We create item-level steering vectors by identifying each single-token word, \(w\), that falls within a given target category and computing the steering vector as \(v_{w,L} = \operatorname{normalize}\!\left(J[L]^\top W_U[w]^\top\right)\), where \(W_U[w]\) is the word's unembedding row vector, and \(J[L]\) maps layer-\(L\) activations to the final residual stream via the J-lens. We obtain an item-level steering vector that broadly targets a category by averaging and then normalising the unit steering vectors for all member tokens in the parent category's token set, \(T\): \(d_{T,L} = \operatorname{normalize}\!\left(\frac{1}{|T|}\sum_{w\in T}v_{w,L}\right)\). To create category-related steering vectors, we use the same category-related labels as reported previously (Fig.~3E) and compute each label's steering vector as \(d_{c,L} = \operatorname{normalize}\!\left(J[L]^\top W_U[c]^\top\right)\), where \(c\) is a single-token category-related label.

In our experiments, we apply a steering vector (\(d_{T,L}\) for item-level steering or \(d_{c,L}\) for category-related steering) during SFT generation, aiming to shift the model into the target category. We first generate an unsteered baseline sequence and then select a target token position at which the following hold: 1) the current item belongs to the same category as the preceding token (the model is clustering); 2) the target category is not among the item's own categories. After identifying the target position, we task the model with continuing the SFT from that position while adding the steering vector to the residual-stream output of block \(L\), such that \(h_{L,t} \leftarrow h_{L,t} + s\,r_L\,d_L\), where \(d_L = d_{T,L}\) for item-level steering and \(d_L = d_{c,L}\) for category-related steering, \(t\) indexes the token position at which steering is applied, the steering coefficient \(s\) controls steering strength, and \(r_L\) is the mean residual norm at that layer, measured once per layer from a fixed reference generation. For all models and experiments, we used an \(s\) of 0.5, except for Qwen2.5-7B, which we found to be easily disrupted by steering; for this model, we used \(s=0.25\).  After applying steering at a target token position, we see if the first concept produced in the continuation belongs to the target category (i.e., is a switch). This experiment was iterated 100 times for each target category.

Across all models, both modes of steering are effective at inducing switching into target categories (Fig. 4B-C), with the probability of switching into a target category for Gemma-9B reaching 0.68, for item-level steering (layer 40), and 0.45, for category-related steering (layer 32). While item-level steering tends to increase in potency up until the final tested layer (Fig. 4B), category-related steering is most potent when applied to the middle or mid-to-late layers of most models (Fig. 4C), supporting the idea that middle LLM layers are more important for representing abstract information \citep{skean2025layer,cheng2025emergence}. These results confirm that concept activations at the item and category level have causal influence on LLM switch decision-making during the SFT. 

\section{Study 2: Generic signatures of switching in activation space}
Our results so far show that the activations of individual concepts anticipate and causally influence switching into specific semantic categories. In Study 2, we ask if  switching behaviour is coupled to \textit{generic} directions in model activation space (i.e., directions that represent switching as such rather than switching into a particular category). If so, this would provide evidence for LLMs representing semantic exploration (switching) versus exploitation (clustering) as generic behaviors.

\subsection{Comparing residual stream activations between clustering and switching} 
We first compare model activations between switching and clustering events. We consider residual stream activations at two candidate positions in relation to each cluster/switch event (Fig. 5A). The \textit{anticipatory} position considers activations at the token position immediately before switching (Fig. 5A). This position tests if there are activation signatures in advance of the model committing to switching. The \textit{event} position considers residual stream states at the position of the switch event itself; this tests if clustering/switching are represented distinctively at the point of emission.

To evaluate the power of discovered directions, we initially generate 100 SFTs per model. Then, in a k-fold (k=5) cross-validation procedure, we take all layer $L$ residual stream representations for each anticipatory or event position of items in our training set and perform Principal Component Analysis (PCA), to reduce their dimensionality to 30. Principal component loadings are used as predictors in L2-regularised logistic regression models (regulation strength=0.1), predicting whether the activations are paired with a switching or clustering event. This procedure is repeated for each fold, scoring the AUC for that layer. While an AUC of 0.50 would indicate that activations predict switching behaviour at chance, an AUC of 1 indicates perfect prediction. Cross-validation is performed such that no category appears in both training and test sets, minimising the influence of semantics. 

As shown in Figure 5B-C, residual stream activations at both anticipatory and event positions are able to predict switching behaviour well above chance (Max AUC for Gemma-9B, anticipatory=0.72, layer 26; event=0.87, layer 16). Interestingly, while anticipatory predictive power is relatively consistent across model layers (Fig. 5B), prediction of switching at the event position is greatest for the middle layers (Fig. 5C). Furthermore, we find that as models produce more items within a given cluster, their activations load progressively more strongly onto the anticipatory direction (Fig. 5D, Table \ref{tab:ramping}, LR, all models : $\chi^2\geq 20.44$, $p<0.0001$); this invites comparison with analogous ramping neural activations in humans during clustering \citep{lundin2023neural}.
\begin{figure*}[t]
  \centering
  \includegraphics[width=1\textwidth]{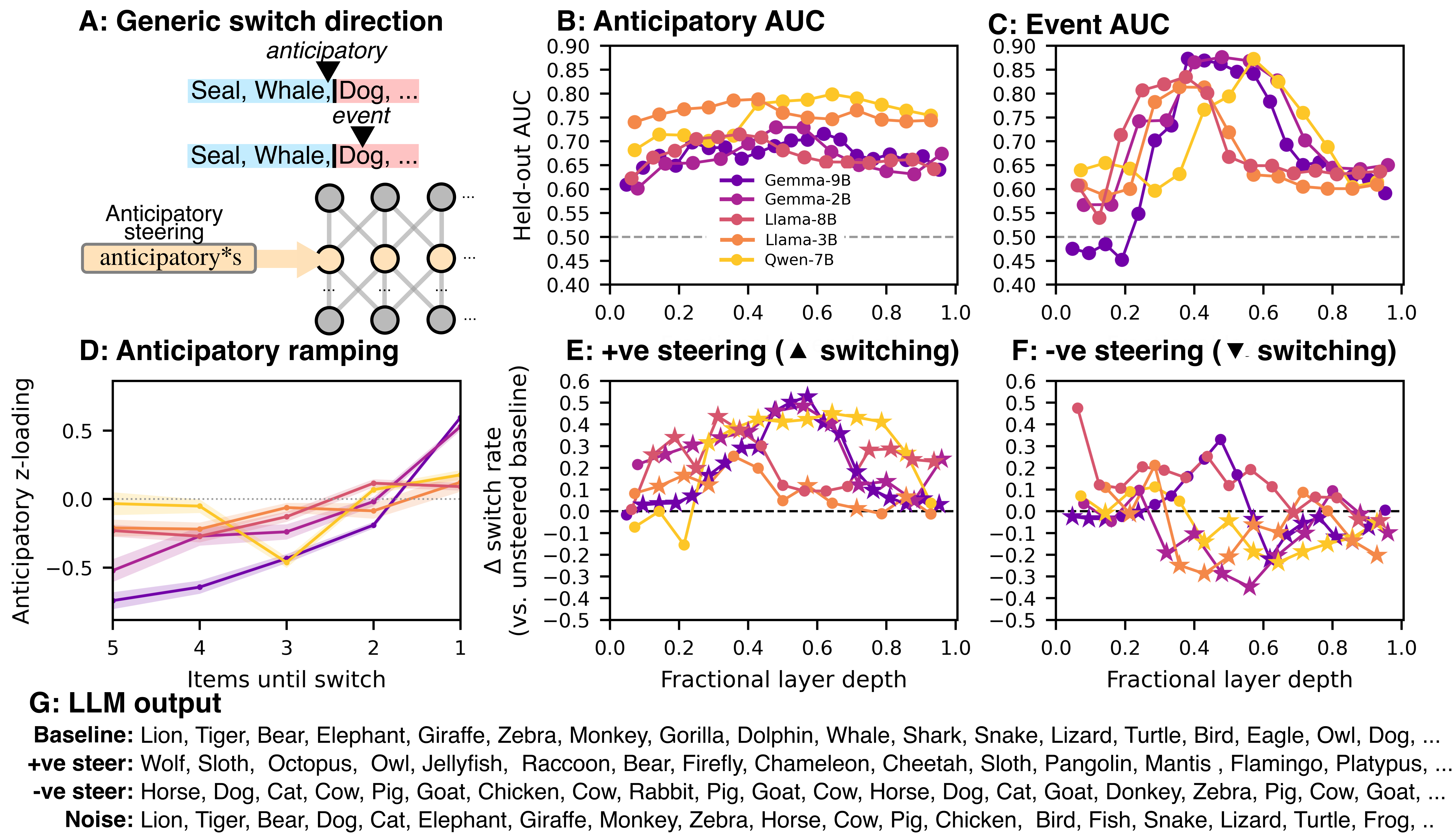}
  \caption{\textbf{Identifying and steering a generic direction for switching in model activation space.} \textbf{A.} Position of anticipatory and event tokens during SFT emission. Bottom neural net: anticipatory steering vectors, weighted by $s$, are used to bias switching. Positive $s$ should bias increased switching and negative $s$, decreased. \textbf{B-C.} AUC from logistic regression models applied across layers and models for predicting switching versus clustering based on anticipatory (\textbf{B}) and event activations (\textbf{C}). \textbf{D.} Layer-averaged loading of model activations onto the anticipatory residual stream direction during clustering, up until switching. Anticipatory activations progressively ramp nearing a switch. Note that Qwen (orange) has the smallest mean cluster size in our experiments (2.4$\pm$2.8 ). \textbf{E-F.} Average effect of positive (\textbf{E}) and negative steering (\textbf{F}) on switching across models and layers, relative to the average switch rate for each model. Points represent means across 100 SFTs. Stars for means indicate significant effects of steering as compared to noise. \textbf{G.} SFT output from Gemma-2-9B under baseline and each perturbation at layer 26.}
  \label{fig:figure_generic}
\end{figure*}
\subsection{Steering semantic exploration-exploitation}
We now ask if these activations have a causal influence on switching by deriving generic “explore/switch" steering vectors for anticipatory positions. If so, then these steering vectors should be able to bias models towards switching or clustering. We generate steering vectors by first fitting logistic regression on the full set of PCA loadings over layer-\(L\) anticipatory residual-stream vectors. We then use the 30 component loadings, \(W \in \mathbb{R}^{D \times 30}\),  in tandem with the logistic regression coefficients, \(\beta \in \mathbb{R}^{30}\), to recover a direction in the original \(D\)-dimensional residual-stream space by projecting the logistic regression's coefficients through the principal component loadings,
\[
d_L = W\beta = \sum_{j=1}^{30} \beta_j w_j,
\]
where \(w_j\) is the \(j\)-th column of \(W\) (the \(j\)-th principal component's loading vector), and normalising to unit length, $\hat{d}_L = \frac{d_L}{\lVert d_L \rVert_2}.
$ The resulting vector, \(\hat{d}_L\), is the generic switching direction at layer \(L\). 

After generating the anticipatory \(\hat{d}_L\), we then use the vector to bias switching behaviour by injecting it into specific residual stream layers during SFT generation: $h_{L,t} \leftarrow h_{L,t} + s\,r_L\,\hat{d}_L,$
where  \(r_L\) is the average residual-stream norm at layer \(L\), measured from an unsteered generation, and $s$ is the steering coefficient, which controls the sign and magnitude of steering. Positive values of $s$ should increase switch rate; negative values of $s$ should decrease the switch rate.  We use \(s \in \{0.1, -0.1\}\) for Qwen-7B and \(s \in \{0.5, -0.5\}\) for all other models.

Furthermore, to test if \textit{any} residual stream perturbation of this magnitude induces changes to the switching rate, we additionally perform experiments in which \(\hat{d}_L\) is substituted for a random direction in the residual stream,  \(\hat{d}_{\mathrm{noise}}\), with its component along \(\hat{d}_L\) removed via Gram--Schmidt orthogonalisation, and injected with the same norm and steering coefficient. For each steering manipulation, we statistically compare the effect of either positive or negative steering to corresponding noise manipulations (See Appendix).

Consistent with the hypothesis that anticipatory activations causally influence switching, we find that steering along this direction is effective at biasing the rate of switching, although the potency of this manipulation was model- and layer-dependent and the effect of positive steering was more consistent than for negative steering (Fig. \ref{fig:figure_generic}E-F, Table \ref{tab:positive_steering}-\ref{tab:negative_steering_stats}, Fisher exact test, all models at most effective layers: $p<0.0001$). For example, positive steering at layer 26 increases the switch rate of Gemma-9B from 0.29 to 0.70; negative steering at the same layer causes switch rate to fall to 0.07. Negative steering was less effective for Llama-3.2-8B (Fig. \ref{fig:figure_generic}F).  These results provide evidence for generic directions in activation space as governing the balance of semantic explore-exploit.

\section{Conclusions}
This study provides evidence that LLMs performing canonical semantic search tasks exhibit distinct representational signatures during exploration and exploitation search regimes. Our results reveal a number of striking parallels in the behaviours and representational dynamics of humans and LLMs. Namely, human SFT findings---elevated response latency during switching \citep{hills2012optimal}, activation of category semantic representations \citep{rissmanWordsNotJust2024,diraniTimeCourseCrossmodal2023}, and neural ramping activation during clustering \citep{lundin2023neural}---are mirrored in LLMs by elevated uncertainty during switching, activation of category-related labels, and ramping activation of switch-associated activations during clustering. Understanding whether these similarities are superficial or reflect deep convergences in the computational processes governing semantic search is a promising avenue for further research.

These findings have several important implications for the use of LLMs. First, if these findings generalise from SFT to narrative generation \citep{leckiePsychosisassociatedDisruptionsSemantic2026}, then activation signatures similar to those identified here could be modulated to control the semantic pace, coherence, and coverage of generated text during routine LLM conversations. Moreover, if exploration-exploitation is similarly represented during more applied search tasks in LLMs (e.g., searching over solution space), it may be possible to steer such searches toward increased or decreased exploration as is useful.

Our study has several limitations and opens up several directions for future research. First, although the SFT allows for tightly controlled examinations of semantic recall dynamics, it is still unclear to what extent the dynamics reported here extend to naturalistic speech or to other search tasks \citep{hills2008search}. Future work could replicate our study in discourse or test the cross-task generality of the switch-associated directions identified in this study.  Second, while generic steering was effective at biasing switching in most models, it is unclear why some models (Llama-3.1-8B) were less susceptible to steering or why the potency of steering was selective across layers and coefficients. Finally, the models used here are relatively small in comparison to present frontier models; it is an open question as to whether these same dynamics emerge in these models.

Overall, this study provides evidence that LLMs search over internal conceptual spaces, representing exploration and exploitation. By extending the semantic foraging framework to artificial intelligences, our work suggests avenues for interpreting, modulating, and enhancing the myriad search tasks performed by LLMs.

\section{AI-use statement}
In this work, we used generative AI tools to generate code (Claude code, Opus models). We have not used generative AI tools for writing or idea generation. LL checked all code manually for correctness and takes full responsibility for the final content of this work, including text, claims or artifacts produced with the aid of generative AI.


\bibliography{LLM_foraging}
\bibliographystyle{iclr2027_conference}

\appendix
\section{Appendix}
\subsection{Switch/cluster designation}
To ensure that switch/cluster designation was robust to an LLM naming a concept outside of our category labels, we omitted verbalised concepts that fell outside of our category labels from cluster-switch designation. However, this was a rare occurence in our experiments (2.4\% of named items across all models pooled, ranging from 0.0\% for Gemma-2-9B to 5.5\% for Qwen2.5-7B). Since some concepts can be multi-word, for experiments in which individual token-level activations were derived, we matched concepts to tokens based upon their first token. Only single-token concepts were used to construct steering vectors. The mean number of items recalled was 36.0$\pm$5.2 for Gemma-2-9B, 41.1$\pm$13.0 for Gemma-2-2B, 64.8$\pm$8.4 for Llama-3.1-8B, 52.1$\pm$11.3 for Llama-3.2-3B, and 39.3$\pm$17.8 for Qwen2.5-7B. Mean cluster size was 3.2$\pm$2.4 for Gemma-2-9B, 2.7$\pm$2.4 for Gemma-2-2B, 2.9$\pm$3.0 for Llama-3.1-8B, 2.2$\pm$2.2 for Llama-3.2-3B, and 2.4$\pm$2.8 for Qwen2.5-7B. 

\subsection{Statistics}
Every significance test treated each seed as one unit of replication, that contributes a single observation at each tested point in statistical tests. For the results presented in Figures 2-3 and Figure 5, independent tests were performed for each model at either each layer/layer block (Fig. 2C-D, Fig. 5D-E) or at each lag value (Fig. 3D, effect averaged over layers). In these tests, we perform linear regression (LR) with Wald's test (Python package: \textit{statsmodels} \cite{seabold2010statsmodels}) to compare the per-seed paired difference in concept activations between clustering and switching (Fig. 2) or category-label activations when switching to a related versus unrelated cluster (Fig. 3). Tests are one-sided (Fig. 3D, Fig. 5E-F) or two-sided, as specified. Where reported, one-sided p-values are multiplied by two. Where tests were performed across multiple layers or lag values for a given model, we protect against multiple testing with Benjamini-Hochberg correction \citep{thissen2002quick}.

To test if the effect of J-space in-category depletion on switching is greater than the effect of depleting the overall pool of animals within a given category depletion we computed the point-biserial correlation between the binary switch outcome and (i) J-space supply and (ii) the raw remaining pool size, using each seed's SFT trials. For interpretability, we divided each model's layers up into five approximately equal depth bins and performed this test at each depth bin, for each model. Within each depth bin, we then tested whether the  J-space depletion correlation was significantly more negative than the raw-depletion correlation, by using the same LR approach described above.

When testing for the effect of anticipatory ramping (Fig. 5D), we performed LRs testing for the association between the number of items remaining in a cluster and the loading onto the anticipatory direction (averaged across all layers). To compare the effect of steering to the effect of stochastic noise on switch rate (Fig. 5E-F), we pooled raw switch/no-switch counts across all seeds and categories at each layer and computed Fisher's exact test comparing the steered condition against a magnitude-matched noise control. The test direction was chosen to match the sign of the steering scale (steer $>$ noise for positive-scale steering, steer $<$ noise for negative-scale steering).

\subsection{Robustness analysis}
To test the robustness of the J-space depletion effect (Fig. 2D) to J-space size, we repeated our experiments on the Gemma models with J-space sizes of 5, and 15 tokens, in addition to the default size of 25 tokens. We did not test sizes larger than 25 since most Troyer categories contain fewer than 25 animals. As shown in Table \ref{tab:j-space_robustness}, our results for Gemma-9B were robust across all tested J-space sizes and our findings for Gemma-2B were robust up until the smallest J-space size of 5. 

\subsection{Noise injection}
We constructed random control steering vectors orthogonal to the
corresponding unit steering direction \(d_L\) as follows:
\[
d_{\mathrm{noise},L}
=
\frac{r - (r^\top d_L)d_L}
{\left\lVert r - (r^\top d_L)d_L \right\rVert_2},
\qquad
r \sim \mathcal{N}(0, I_D),
\]
where \(D\) is the residual-stream dimension. We injected these control
vectors using the same layer-specific residual norm \(r_L\) and steering
coefficient \(s\) as for the corresponding constructed steering vectors,
giving an identical perturbation norm of \(\lvert s\rvert r_L\).

\subsection{Supplementary figures}
\setcounter{figure}{0}
\setcounter{table}{0}
\renewcommand{\thefigure}{S\arabic{figure}}
\renewcommand{\thetable}{S\arabic{table}}
\begin{figure}[H]
  \centering
  \includegraphics[width=0.7\textwidth]{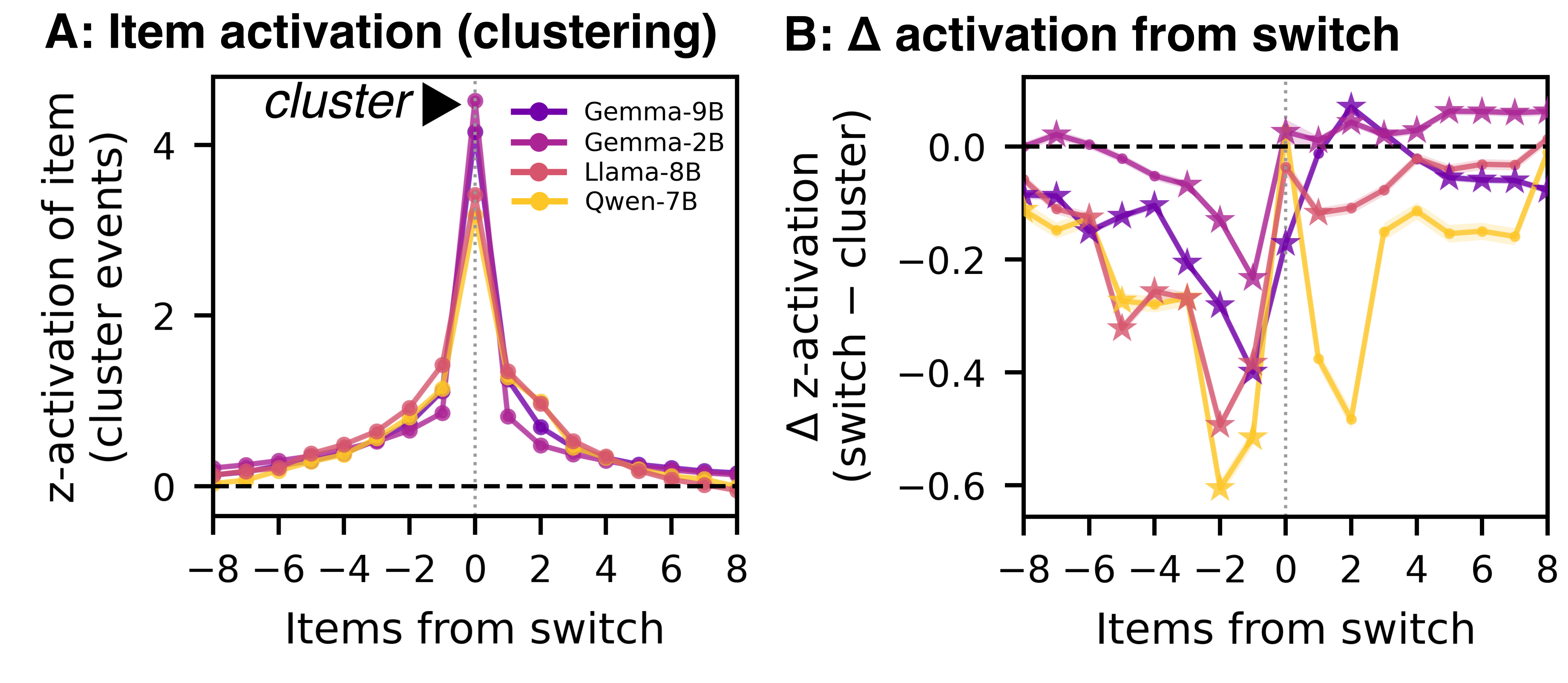}
   \label{fig:suppjlenss}
  \caption{\textbf{A.} Z-scored J-lens activations for items participating in clustering, measured in advance (negative lag) and after (positive lag) their emission (lag=0). Points represent means across seeds and layers and shaded areas, standard errors. \textbf{C.}  $\Delta$ in Z-scored J-lens activations for switching minus clustering. Points represent means and shaded areas, standard errors (note that standard errors are very narrow). Stars indicate significant comparisons between switch- and cluster- associated activations.}
\end{figure}
\subsection{Supplementary tables}
\begin{table}[H]
\caption{Statistics comparing J-lens activations of items participating in clustering or switching across layers (Fig. 2C).}
\label{tab:jlens}
\small
\begin{tabular}{|l|l|l|l|l|l|l|}\hline
model    & layer & beta   & F        & df & p& sig   \\\hline
Gemma-9B & 2     & -0.187 & 242.080  & 1  & 2.84E-28          & TRUE  \\\hline
Gemma-9B & 4     & -0.244 & 263.261  & 1  & 1.51E-29          & TRUE  \\\hline
Gemma-9B & 6     & -0.176 & 130.854  & 1  & 8.64E-20          & TRUE  \\\hline
Gemma-9B & 8     & -0.259 & 544.888  & 1  & 8.83E-42          & TRUE  \\\hline
Gemma-9B & 10    & -0.415 & 951.408  & 1  & 4.04E-52          & TRUE  \\\hline
Gemma-9B & 12    & -0.411 & 1011.960 & 1  & 2.94E-53          & TRUE  \\\hline
Gemma-9B & 14    & -0.389 & 647.727  & 1  & 6.96E-45          & TRUE  \\\hline
Gemma-9B & 16    & -0.304 & 310.668  & 1  & 3.59E-32          & TRUE  \\\hline
Gemma-9B & 18    & -0.155 & 72.541   & 1  & 1.85E-13          & TRUE  \\\hline
Gemma-9B & 20    & -0.287 & 457.983  & 1  & 1.07E-38          & TRUE  \\\hline
Gemma-9B & 22    & -0.502 & 642.140  & 1  & 9.09E-45          & TRUE  \\\hline
Gemma-9B & 24    & -0.354 & 223.009  & 1  & 4.69E-27          & TRUE  \\\hline
Gemma-9B & 26    & -0.463 & 366.078  & 1  & 7.63E-35          & TRUE  \\\hline
Gemma-9B & 28    & -0.490 & 821.303  & 1  & 2.48E-49          & TRUE  \\\hline
Gemma-9B & 30    & -0.567 & 1453.269 & 1  & 2.80E-60          & TRUE  \\\hline
Gemma-9B & 32    & -0.795 & 2279.001 & 1  & 7.49E-69          & TRUE  \\\hline
Gemma-9B & 34    & -0.756 & 1662.956 & 1  & 1.05E-62          & TRUE  \\\hline
Gemma-9B & 36    & -0.793 & 1611.530 & 1  & 3.05E-62          & TRUE  \\\hline
Gemma-9B & 38    & -0.596 & 1081.406 & 1  & 1.75E-54          & TRUE  \\\hline
Gemma-9B & 40    & -0.377 & 354.221  & 1  & 2.55E-34          & TRUE  \\\hline
Gemma-2B & 2     & 0.077  & 64.162   & 1  & 2.90E-12          & TRUE  \\\hline
Gemma-2B & 4     & 0.018  & 3.132    & 1  & 8.71E-2 & FALSE \\\hline
Gemma-2B & 6     & -0.004 & 0.115    & 1  & 7.35E- & FALSE \\\hline
Gemma-2B & 8     & -0.098 & 98.177   & 1  & 2.87E-16          & TRUE  \\\hline
Gemma-2B & 10    & -0.080 & 65.834   & 1  & 1.93E-12          & TRUE  \\\hline
Gemma-2B & 12    & -0.246 & 369.993  & 1  & 9.65E-35          & TRUE  \\\hline
Gemma-2B & 14    & -0.219 & 267.830  & 1  & 1.50E-29          & TRUE  \\\hline
Gemma-2B & 16    & -0.406 & 661.402  & 1  & 1.38E-44          & TRUE  \\\hline
Gemma-2B & 18    & -0.427 & 847.678  & 1  & 5.84E-49          & TRUE  \\\hline
Gemma-2B & 20    & -0.363 & 387.246  & 1  & 1.96E-35          & TRUE  \\\hline
Gemma-2B & 22    & -0.369 & 511.008  & 1  & 4.66E-40          & TRUE  \\\hline
Gemma-2B & 24    & -0.500 & 491.088  & 1  & 1.79E-39          & TRUE  \\\hline
Qwen-7B  & 2     & -0.042 & 0.992    & 1  & 2.23E-1 & FALSE \\\hline
Qwen-7B  & 4     & -0.179 & 13.834   & 1  & 4.90E-4 & TRUE  \\\hline
Qwen-7B  & 6     & -0.153 & 16.619   & 1  & 1.83E-4 & TRUE  \\\hline
Qwen-7B  & 8     & -0.114 & 16.094   & 1  & 2.07E-4 & TRUE  \\\hline
Qwen-7B  & 10    & -0.375 & 105.316  & 1  & 1.53E-13          & TRUE  \\\hline
Qwen-7B  & 12    & -0.358 & 81.342   & 1  & 4.01E-12          & TRUE  \\\hline
Qwen-7B  & 14    & -0.349 & 85.948   & 1  & 2.06E-12          & TRUE  \\\hline
Qwen-7B  & 16    & -0.464 & 72.395   & 1  & 2.25E-11          & TRUE  \\\hline
Qwen-7B  & 18    & -0.379 & 54.639   & 1  & 1.39E-09          & TRUE  \\\hline
Qwen-7B  & 20    & -0.424 & 46.903   & 1  & 9.77E-09          & TRUE  \\\hline
Qwen-7B  & 22    & -0.469 & 32.364   & 1  & 7.17E-07          & TRUE  \\\hline
Qwen-7B  & 24    & -0.457 & 18.501   & 1  & 9.53E-05          & TRUE  \\\hline
Qwen-7B  & 26    & -0.617 & 92.662   & 1  & 8.12E-13          & TRUE  \\\hline
Llama-8B & 2     & 0.030  & 1.293    & 1  & 2.58E-1 & FALSE \\\hline
Llama-8B & 4     & 0.095  & 22.290   & 1  & 9.11E-06          & TRUE  \\\hline
Llama-8B & 6     & -0.050 & 3.266    & 1  & 7.91E-2  & FALSE \\\hline
Llama-8B & 8     & -0.228 & 55.423   & 1  & 4.98E-11          & TRUE  \\\hline
Llama-8B & 10    & -0.249 & 89.535   & 1  & 2.92E-15          & TRUE  \\\hline
Llama-8B & 12    & -0.216 & 88.239   & 1  & 3.73E-15          & TRUE  \\\hline
Llama-8B & 14    & -0.209 & 121.252  & 1  & 1.49E-18          & TRUE  \\\hline
Llama-8B & 16    & -0.351 & 147.041  & 1  & 8.17E-21          & TRUE  \\\hline
Llama-8B & 18    & -0.484 & 327.454  & 1  & 5.57E-32          & TRUE  \\\hline
Llama-8B & 20    & -0.480 & 375.362  & 1  & 6.14E-34          & TRUE  \\\hline
Llama-8B & 22    & -0.503 & 287.428  & 1  & 3.45E-30          & TRUE  \\\hline
Llama-8B & 24    & -0.497 & 229.206  & 1  & 6.83E-27          & TRUE  \\\hline
Llama-8B & 26    & -0.614 & 305.371  & 1  & 5.00E-31          & TRUE  \\\hline
Llama-8B & 28    & -0.719 & 243.439  & 1  & 1.02E-27          & TRUE  \\\hline
Llama-8B & 30    & -0.571 & 132.399  & 1  & 1.47E-19          & TRUE 
\\ \hline\end{tabular}
\end{table}

{\footnotesize
\begin{longtable}{|l|l|l|l|l|l|l|}
\caption{Statistics comparing J-lens activations of items participating in clustering or switching across lag values from their emission, averaged over layers (Fig. S1B).}
\label{tab:J-lens-activations-lag}\\
\hline
model    & item lag& beta   & F        & df & p        & sig   \\\hline
Gemma-9B & -10       & 0.009  & 0.245    & 1  & 6.51E-01 & FALSE \\\hline
Gemma-9B & -9        & 0.016  & 0.768    & 1  & 4.47E-01 & FALSE \\\hline
Gemma-9B & -8        & -0.066 & 11.302   & 1  & 1.93E-03 & TRUE  \\\hline
Gemma-9B & -7        & -0.079 & 16.463   & 1  & 2.08E-04 & TRUE  \\\hline
Gemma-9B & -6        & -0.148 & 59.068   & 1  & 4.72E-11 & TRUE  \\\hline
Gemma-9B & -5        & -0.124 & 42.628   & 1  & 8.58E-09 & TRUE  \\\hline
Gemma-9B & -4        & -0.097 & 43.818   & 1  & 6.56E-09 & TRUE  \\\hline
Gemma-9B & -3        & -0.201 & 228.004  & 1  & 1.37E-26 & TRUE  \\\hline
Gemma-9B & -2        & -0.274 & 470.084  & 1  & 2.33E-38 & TRUE  \\\hline
Gemma-9B & -1        & -0.394 & 2671.882 & 1  & 4.05E-72 & TRUE  \\\hline
Gemma-9B & 0         & -0.168 & 107.145  & 1  & 9.86E-17 & TRUE  \\\hline
Gemma-9B & 1         & -0.004 & 0.206    & 1  & 6.51E-01 & FALSE \\\hline
Gemma-9B & 2         & 0.085  & 36.602   & 1  & 6.80E-08 & TRUE  \\\hline
Gemma-9B & 3         & 0.037  & 4.126    & 1  & 5.55E-02 & FALSE \\\hline
Gemma-9B & 4         & -0.014 & 0.637    & 1  & 4.71E-01 & FALSE \\\hline
Gemma-9B & 5         & -0.044 & 7.18     & 1  & 1.13E-02 & TRUE  \\\hline
Gemma-9B & 6         & -0.052 & 10.794   & 1  & 2.28E-03 & TRUE  \\\hline
Gemma-9B & 7         & -0.055 & 8.933    & 1  & 5.30E-03 & TRUE  \\\hline
Gemma-9B & 8         & -0.076 & 15.738   & 1  & 2.63E-04 & TRUE  \\\hline
Gemma-9B & 9         & -0.054 & 8.059    & 1  & 7.69E-03 & TRUE  \\\hline
Gemma-9B & 10        & -0.083 & 16.534   & 1  & 2.08E-04 & TRUE  \\\hline
Gemma-2B & -10       & 0.032  & 3.766    & 1  & 6.82E-02 & FALSE \\\hline
Gemma-2B & -9        & 0.024  & 1.928    & 1  & 1.78E-01 & FALSE \\\hline
Gemma-2B & -8        & 0.023  & 1.918    & 1  & 1.78E-01 & FALSE \\\hline
Gemma-2B & -7        & 0.041  & 8.99     & 1  & 4.82E-03 & TRUE  \\\hline
Gemma-2B & -6        & 0.026  & 2.29     & 1  & 1.56E-01 & FALSE \\\hline
Gemma-2B & -5        & 0.004  & 0.068    & 1  & 7.94E-01 & FALSE \\\hline
Gemma-2B & -4        & -0.028 & 4.318    & 1  & 5.29E-02 & FALSE \\\hline
Gemma-2B & -3        & -0.043 & 13.571   & 1  & 5.65E-04 & TRUE  \\\hline
Gemma-2B & -2        & -0.106 & 145.214  & 1  & 4.92E-20 & TRUE  \\\hline
Gemma-2B & -1        & -0.208 & 775.117  & 1  & 5.13E-47 & TRUE  \\\hline
Gemma-2B & 0         & 0.052  & 17.543   & 1  & 1.17E-04 & TRUE  \\\hline
Gemma-2B & 1         & 0.041  & 26.075   & 1  & 4.24E-06 & TRUE  \\\hline
Gemma-2B & 2         & 0.078  & 41.449   & 1  & 3.15E-08 & TRUE  \\\hline
Gemma-2B & 3         & 0.056  & 18.202   & 1  & 9.63E-05 & TRUE  \\\hline
Gemma-2B & 4         & 0.066  & 19.887   & 1  & 5.11E-05 & TRUE  \\\hline
Gemma-2B & 5         & 0.095  & 34.362   & 1  & 3.24E-07 & TRUE  \\\hline
Gemma-2B & 6         & 0.089  & 29.852   & 1  & 1.24E-06 & TRUE  \\\hline
Gemma-2B & 7         & 0.08   & 32.369   & 1  & 5.56E-07 & TRUE  \\\hline
Gemma-2B & 8         & 0.08   & 15.309   & 1  & 2.95E-04 & TRUE  \\\hline
Gemma-2B & 9         & 0.082  & 13.864   & 1  & 5.30E-04 & TRUE  \\\hline
Gemma-2B & 10        & 0.123  & 28.293   & 1  & 1.98E-06 & TRUE  \\\hline
Qwen-7B  & -10       & -0.098 & 2.619    & 1  & 2.32E-01 & FALSE \\\hline
Qwen-7B  & -9        & -0.043 & 0.808    & 1  & 4.88E-01 & FALSE \\\hline
Qwen-7B  & -8        & -0.143 & 7.433    & 1  & 4.39E-02 & TRUE  \\\hline
Qwen-7B  & -7        & 0      & 0        & 1  & 9.96E-01 & FALSE \\\hline
Qwen-7B  & -6        & -0.123 & 2.9      & 1  & 2.20E-01 & FALSE \\\hline
Qwen-7B  & -5        & -0.171 & 8.523    & 1  & 3.51E-02 & TRUE  \\\hline
Qwen-7B  & -4        & -0.122 & 4.627    & 1  & 1.22E-01 & FALSE \\\hline
Qwen-7B  & -3        & -0.153 & 7.005    & 1  & 4.39E-02 & TRUE  \\\hline
Qwen-7B  & -2        & -0.366 & 52.617   & 1  & 1.15E-08 & TRUE  \\\hline
Qwen-7B  & -1        & -0.309 & 72.551   & 1  & 1.76E-10 & TRUE  \\\hline
Qwen-7B  & 0         & -0.074 & 2.47     & 1  & 2.32E-01 & FALSE \\\hline
Qwen-7B  & 1         & -0.031 & 0.733    & 1  & 4.88E-01 & FALSE \\\hline
Qwen-7B  & 2         & -0.096 & 4.384    & 1  & 1.22E-01 & FALSE \\\hline
Qwen-7B  & 3         & 0.058  & 0.827    & 1  & 4.88E-01 & FALSE \\\hline
Qwen-7B  & 4         & 0.053  & 1.111    & 1  & 4.78E-01 & FALSE \\\hline
Qwen-7B  & 5         & -0.003 & 0.002    & 1  & 9.96E-01 & FALSE \\\hline
Qwen-7B  & 6         & 0.036  & 0.282    & 1  & 6.60E-01 & FALSE \\\hline
Qwen-7B  & 7         & -0.044 & 0.529    & 1  & 5.48E-01 & FALSE \\\hline
Qwen-7B  & 8         & 0.104  & 1.953    & 1  & 2.93E-01 & FALSE \\\hline
Qwen-7B  & 9         & -0.116 & 3.291    & 1  & 1.97E-01 & FALSE \\\hline
Qwen-7B  & 10        & 0.063  & 0.991    & 1  & 4.86E-01 & FALSE \\\hline
Llama-8B & -10       & 0.043  & 1.805    & 1  & 2.94E-01 & FALSE \\\hline
Llama-8B & -9        & -0.024 & 0.799    & 1  & 5.23E-01 & FALSE \\\hline
Llama-8B & -8        & -0.045 & 2.554    & 1  & 2.16E-01 & FALSE \\\hline
Llama-8B & -7        & -0.068 & 5.082    & 1  & 6.94E-02 & FALSE \\\hline
Llama-8B & -6        & -0.087 & 9.324    & 1  & 8.76E-03 & TRUE  \\\hline
Llama-8B & -5        & -0.285 & 86.335   & 1  & 3.18E-14 & TRUE  \\\hline
Llama-8B & -4        & -0.191 & 33.069   & 1  & 4.33E-07 & TRUE  \\\hline
Llama-8B & -3        & -0.225 & 68.183   & 1  & 3.99E-12 & TRUE  \\\hline
Llama-8B & -2        & -0.445 & 191.017  & 1  & 1.24E-23 & TRUE  \\\hline
Llama-8B & -1        & -0.358 & 413.825  & 1  & 1.91E-35 & TRUE  \\\hline
Llama-8B & 0         & -0.017 & 0.367    & 1  & 6.37E-01 & FALSE \\\hline
Llama-8B & 1         & -0.068 & 16.039   & 1  & 4.26E-04 & TRUE  \\\hline
Llama-8B & 2         & -0.046 & 3.626    & 1  & 1.26E-01 & FALSE \\\hline
Llama-8B & 3         & -0.005 & 0.034    & 1  & 8.66E-01 & FALSE \\\hline
Llama-8B & 4         & 0.04   & 2.385    & 1  & 2.20E-01 & FALSE \\\hline
Llama-8B & 5         & 0.018  & 0.41     & 1  & 6.37E-01 & FALSE \\\hline
Llama-8B & 6         & 0.012  & 0.207    & 1  & 7.19E-01 & FALSE \\\hline
Llama-8B & 7         & 0.018  & 0.472    & 1  & 6.37E-01 & FALSE \\\hline
Llama-8B & 8         & 0.036  & 1.288    & 1  & 3.89E-01 & FALSE \\\hline
Llama-8B & 9         & -0.004 & 0.029    & 1  & 8.66E-01 & FALSE \\\hline
Llama-8B & 10        & 0.072  & 3.618    & 1  & 1.26E-01 & FALSE
\end{longtable}
}
\newpage

\begin{table}[H]
\caption{Statistics to accompany Fig. 2D. Pooled correlation coefficients (r) are tested against 0 with one-sample t-tests.}
\label{tab:j-spacedepletion_base}
\small
\begin{tabular}{|l|l|l|l|l|l|l|}\hline
model    & depth range & r      & t       & df & p        & sig   \\\hline
Gemma-9B & 0.00-0.20   & -0.367 & -30.358 & 1  & 1.47E-51 & TRUE  \\\hline
Gemma-9B & 0.20-0.40   & -0.402 & -26.220 & 1  & 2.99E-46 & TRUE  \\\hline
Gemma-9B & 0.40-0.60   & -0.390 & -31.876 & 1  & 3.66E-53 & TRUE  \\\hline
Gemma-9B & 0.60-0.80   & -0.362 & -17.728 & 1  & 1.74E-32 & TRUE  \\\hline
Gemma-9B & 0.80-1.00   & -0.306 & -27.619 & 1  & 4.29E-48 & TRUE  \\\hline
Gemma-2B & 0.00-0.20   & -0.336 & -18.451 & 1  & 1.19E-33 & TRUE  \\\hline
Gemma-2B & 0.20-0.40   & -0.360 & -21.623 & 1  & 1.10E-38 & TRUE  \\\hline
Gemma-2B & 0.40-0.60   & -0.341 & -19.582 & 1  & 1.50E-35 & TRUE  \\\hline
Gemma-2B & 0.60-0.80   & -0.376 & -21.293 & 1  & 2.55E-38 & TRUE  \\\hline
Gemma-2B & 0.80-1.00   & -0.385 & -23.401 & 1  & 3.32E-41 & TRUE  \\\hline
Qwen-7B  & 0.00-0.20   & -0.326 & -1.651  & 1  & 1.05E-01 & FALSE \\\hline
Qwen-7B  & 0.20-0.40   & -0.355 & -4.014  & 1  & 3.27E-04 & TRUE  \\\hline
Qwen-7B  & 0.40-0.60   & -0.332 & -4.046  & 1  & 3.27E-04 & TRUE  \\\hline
Qwen-7B  & 0.60-0.80   & -0.317 & -8.427  & 1  & 1.57E-10 & TRUE  \\\hline
Qwen-7B  & 0.80-1.00   & -0.380 & -1.905  & 1  & 7.80E-02 & FALSE \\\hline
Llama-8B & 0.00-0.20   & -0.377 & -18.379 & 1  & 4.83E-33 & TRUE  \\\hline
Llama-8B & 0.20-0.40   & -0.387 & -18.563 & 1  & 2.87E-33 & TRUE  \\\hline
Llama-8B & 0.40-0.60   & -0.405 & -18.688 & 1  & 2.31E-33 & TRUE  \\\hline
Llama-8B & 0.60-0.80   & -0.425 & -19.329 & 1  & 2.73E-34 & TRUE  \\\hline
Llama-8B & 0.80-1.00   & -0.500 & -22.711 & 1  & 1.81E-39 & TRUE 
\\ \hline\end{tabular}
\end{table}
\begin{table}[H]
\caption{Statistical comparisons testing the effect of J-space in-category depletion on switching probability, as compared to depletion of the overall category pool.}
\label{tab:j-space_supply}
\begin{tabular}{|l|l|l|l|l|l|l|}\hline
model    & depth range& beta   & F       & df & p        & sig   \\\hline
Gemma-9B & 0.00-0.20    & -0.091 & 82.890  & 1  & 1.64E-14 & TRUE  \\\hline
Gemma-9B & 0.20-0.40    & -0.137 & 131.235 & 1  & 1.89E-19 & TRUE  \\\hline
Gemma-9B & 0.40-0.60    & -0.140 & 202.842 & 1  & 5.26E-25 & TRUE  \\\hline
Gemma-9B & 0.60-0.80    & -0.059 & 24.841  & 1  & 3.31E-06 & TRUE  \\\hline
Gemma-9B & 0.80-1.00    & -0.016 & 1.987   & 1  & 1.62E-01 & FALSE \\\hline
Gemma-2B & 0.00-0.20    & -0.039 & 3.717   & 1  & 9.46E-02 & FALSE \\\hline
Gemma-2B & 0.20-0.40    & -0.028 & 1.648   & 1  & 2.33E-01 & FALSE \\\hline
Gemma-2B & 0.40-0.60    & -0.027 & 1.441   & 1  & 2.33E-01 & FALSE \\\hline
Gemma-2B & 0.60-0.80    & -0.049 & 4.348   & 1  & 9.46E-02 & FALSE \\\hline
Gemma-2B & 0.80-1.00    & -0.080 & 19.940  & 1  & 1.07E-04 & TRUE  \\\hline
Qwen-7B  & 0.00-0.20    & 0.027  & 0.503   & 1  & 1.00E+00 & FALSE \\\hline
Qwen-7B  & 0.20-0.40    & -0.067 & 5.371   & 1  & 6.13E-02 & FALSE \\\hline
Qwen-7B  & 0.40-0.60    & -0.076 & 6.139   & 1  & 6.13E-02 & FALSE \\\hline
Qwen-7B  & 0.60-0.80    & -0.064 & 2.781   & 1  & 1.69E-01 & FALSE \\\hline
Qwen-7B  & 0.80-1.00    & -0.038 & 1.619   & 1  & 2.61E-01 & FALSE \\\hline
Llama-8B & 0.00-0.20    & 0.038  & 16.786  & 1  & 1.00E+00 & FALSE \\\hline
Llama-8B & 0.20-0.40    & 0.026  & 4.100   & 1  & 1.00E+00 & FALSE \\\hline
Llama-8B & 0.40-0.60    & 0.010  & 0.699   & 1  & 1.00E+00 & FALSE \\\hline
Llama-8B & 0.60-0.80    & -0.012 & 0.648   & 1  & 1.00E+00 & FALSE \\\hline
Llama-8B & 0.80-1.00    & -0.062 & 21.517  & 1  & 5.60E-05 & TRUE 
\\ \hline\end{tabular}
\end{table}
\begin{table}[H]
\caption{Robustness analyses testing the effect of J-space size on the comparisons reported in \ref{tab:j-space_supply} for Gemma models. Sizes of 5, 15, and 25 were tested.}
\label{tab:j-space_robustness}
\small
\begin{tabular}{|l|l|l|l|l|l|l|l|}\hline
model    & J-space size & depth range & beta   & F       & df & p        & sig   \\\hline
Gemma-9B & 5            & 0.00-0.20   & 0.033  & 4.530   & 1  & 1.00E+00 & FALSE \\\hline
Gemma-9B & 5            & 0.20-0.40   & -0.061 & 15.266  & 1  & 2.85E-04 & TRUE  \\\hline
Gemma-9B & 5            & 0.40-0.60   & -0.060 & 16.248  & 1  & 2.73E-04 & TRUE  \\\hline
Gemma-9B & 5            & 0.60-0.80   & -0.069 & 17.788  & 1  & 2.73E-04 & TRUE  \\\hline
Gemma-9B & 5            & 0.80-1.00   & 0.064  & 16.907  & 1  & 1.00E+00 & FALSE \\\hline
Gemma-9B & 15           & 0.00-0.20   & -0.043 & 16.140  & 1  & 1.43E-04 & TRUE  \\\hline
Gemma-9B & 15           & 0.20-0.40   & -0.134 & 129.976 & 1  & 2.48E-19 & TRUE  \\\hline
Gemma-9B & 15           & 0.40-0.60   & -0.146 & 217.233 & 1  & 5.19E-26 & TRUE  \\\hline
Gemma-9B & 15           & 0.60-0.80   & -0.105 & 84.051  & 1  & 1.19E-14 & TRUE  \\\hline
Gemma-9B & 15           & 0.80-1.00   & -0.011 & 1.163   & 1  & 2.83E-01 & FALSE \\\hline
Gemma-9B & 25 (default) & 0.00-0.20   & -0.091 & 82.890  & 1  & 1.64E-14 & TRUE  \\\hline
Gemma-9B & 25 (default) & 0.20-0.40   & -0.137 & 131.235 & 1  & 1.89E-19 & TRUE  \\\hline
Gemma-9B & 25 (default) & 0.40-0.60   & -0.140 & 202.842 & 1  & 5.26E-25 & TRUE  \\\hline
Gemma-9B & 25 (default) & 0.60-0.80   & -0.059 & 24.841  & 1  & 3.31E-06 & TRUE  \\\hline
Gemma-9B & 25 (default) & 0.80-1.00   & -0.016 & 1.987   & 1  & 1.62E-01 & FALSE \\\hline
Gemma-2B & 5            & 0.00-0.20   & 0.099  & 21.880  & 1  & 1.00E+00 & FALSE \\\hline
Gemma-2B & 5            & 0.20-0.40   & 0.033  & 2.394   & 1  & 1.00E+00 & FALSE \\\hline
Gemma-2B & 5            & 0.40-0.60   & 0.049  & 5.200   & 1  & 1.00E+00 & FALSE \\\hline
Gemma-2B & 5            & 0.60-0.80   & 0.065  & 10.575  & 1  & 1.00E+00 & FALSE \\\hline
Gemma-2B & 5            & 0.80-1.00   & 0.118  & 27.536  & 1  & 1.00E+00 & FALSE \\\hline
Gemma-2B & 15           & 0.00-0.20   & -0.031 & 3.479   & 1  & 6.51E-02 & FALSE \\\hline
Gemma-2B & 15           & 0.20-0.40   & -0.062 & 12.812  & 1  & 1.34E-03 & TRUE  \\\hline
Gemma-2B & 15           & 0.40-0.60   & -0.054 & 9.813   & 1  & 3.14E-03 & TRUE  \\\hline
Gemma-2B & 15           & 0.60-0.80   & -0.054 & 9.620   & 1  & 3.14E-03 & TRUE  \\\hline
Gemma-2B & 15           & 0.80-1.00   & -0.068 & 16.848  & 1  & 4.19E-04 & TRUE  \\\hline
Gemma-2B & 25 (default) & 0.00-0.20   & -0.039 & 3.717   & 1  & 9.46E-02 & FALSE \\\hline
Gemma-2B & 25 (default) & 0.20-0.40   & -0.028 & 1.648   & 1  & 2.33E-01 & FALSE \\\hline
Gemma-2B & 25 (default) & 0.40-0.60   & -0.027 & 1.441   & 1  & 2.33E-01 & FALSE \\\hline
Gemma-2B & 25 (default) & 0.60-0.80   & -0.049 & 4.348   & 1  & 9.46E-02 & FALSE \\\hline
Gemma-2B & 25 (default) & 0.80-1.00   & -0.080 & 19.940  & 1  & 1.07E-04 & TRUE 
\\ \hline\end{tabular}
\end{table}

\begin{table}[H]
\caption{Statistics for Figure 3D, comparing the activation of category-related labels before and after switching into a target category, versus an unrelated category}
\label{tab:category_stats}
\small
\begin{tabular}{|l|l|l|l|l|l|l|}\hline
model    & item & beta   & F         & df & p         & sig   \\\hline
Gemma-9B & -6   & 0.107  & 23.676    & 1  & 4.31E-06  & TRUE  \\\hline
Gemma-9B & -5   & 0.123  & 61.405    & 1  & 6.35E-12  & TRUE  \\\hline
Gemma-9B & -4   & 0.154  & 71.636    & 1  & 3.13E-13  & TRUE  \\\hline
Gemma-9B & -3   & 0.113  & 41.314    & 1  & 4.97E-09  & TRUE  \\\hline
Gemma-9B & -2   & 0.266  & 165.96    & 1  & 1.12E-22  & TRUE  \\\hline
Gemma-9B & -1   & 0.158  & 344.957   & 1  & 1.08E-33  & TRUE  \\\hline
Gemma-9B & 0    & 1.712  & 16928.534 & 1  & 2.29E-111 & TRUE  \\\hline
Gemma-9B & 1    & 0.6    & 3100.687  & 1  & 1.01E-75  & TRUE  \\\hline
Gemma-9B & 2    & 0.548  & 1819.107  & 1  & 6.81E-65  & TRUE  \\\hline
Gemma-9B & 3    & 0.466  & 896.722   & 1  & 6.51E-51  & TRUE  \\\hline
Gemma-9B & 4    & 0.363  & 572.918   & 1  & 1.53E-42  & TRUE  \\\hline
Gemma-9B & 5    & 0.271  & 178.355   & 1  & 1.32E-23  & TRUE  \\\hline
Gemma-9B & 6    & 0.193  & 73.819    & 1  & 1.84E-13  & TRUE  \\\hline
Gemma-2B & -6   & 0.03   & 3.003     & 1  & 8.62E-02  & FALSE \\\hline
Gemma-2B & -5   & 0.05   & 8.545     & 1  & 4.66E-03  & TRUE  \\\hline
Gemma-2B & -4   & 0.069  & 21.05     & 1  & 1.72E-05  & TRUE  \\\hline
Gemma-2B & -3   & 0.087  & 31.561    & 1  & 2.62E-07  & TRUE  \\\hline
Gemma-2B & -2   & 0.119  & 68.154    & 1  & 1.34E-12  & TRUE  \\\hline
Gemma-2B & -1   & 0.117  & 172.203   & 1  & 6.81E-23  & TRUE  \\\hline
Gemma-2B & 0    & 1.779  & 6370.834  & 1  & 7.28E-90  & TRUE  \\\hline
Gemma-2B & 1    & 0.355  & 1894.557  & 1  & 4.24E-65  & TRUE  \\\hline
Gemma-2B & 2    & 0.276  & 486.139   & 1  & 3.91E-39  & TRUE  \\\hline
Gemma-2B & 3    & 0.202  & 207.11    & 1  & 2.14E-25  & TRUE  \\\hline
Gemma-2B & 4    & 0.167  & 130.421   & 1  & 2.25E-19  & TRUE  \\\hline
Gemma-2B & 5    & 0.123  & 47.62     & 1  & 8.37E-10  & TRUE  \\\hline
Gemma-2B & 6    & 0.065  & 10.149    & 1  & 2.29E-03  & TRUE  \\\hline
Qwen-7B  & -6   & 0.208  & 20.058    & 1  & 5.09E-05  & TRUE  \\\hline
Qwen-7B  & -5   & 0.056  & 1.574     & 1  & 3.08E-01  & FALSE \\\hline
Qwen-7B  & -4   & -0.162 & 8.396     & 1  & 1.00E+00  & FALSE \\\hline
Qwen-7B  & -3   & 0.235  & 36.128    & 1  & 1.50E-07  & TRUE  \\\hline
Qwen-7B  & -2   & -0.24  & 11.925    & 1  & 1.00E+00  & FALSE \\\hline
Qwen-7B  & -1   & -0.166 & 6.875     & 1  & 1.00E+00  & FALSE \\\hline
Qwen-7B  & 0    & 1.1    & 848.435   & 1  & 1.47E-44  & TRUE  \\\hline
Qwen-7B  & 1    & 0.135  & 5.667     & 1  & 3.18E-02  & TRUE  \\\hline
Qwen-7B  & 2    & 0.205  & 7.36      & 1  & 1.50E-02  & TRUE  \\\hline
Qwen-7B  & 3    & 0.353  & 24.876    & 1  & 8.75E-06  & TRUE  \\\hline
Qwen-7B  & 4    & 0.516  & 87.216    & 1  & 9.91E-14  & TRUE  \\\hline
Qwen-7B  & 5    & 0.315  & 38.556    & 1  & 9.02E-08  & TRUE  \\\hline
Qwen-7B  & 6    & 0.045  & 0.487     & 1  & 6.33E-01  & FALSE \\\hline
Llama-8B & -6   & -0.021 & 0.591     & 1  & 1.00E+00  & FALSE \\\hline
Llama-8B & -5   & 0.07   & 5.649     & 1  & 2.10E-02  & TRUE  \\\hline
Llama-8B & -4   & 0.102  & 12.769    & 1  & 6.46E-04  & TRUE  \\\hline
Llama-8B & -3   & 0.155  & 39.192    & 1  & 1.61E-08  & TRUE  \\\hline
Llama-8B & -2   & 0.272  & 127.172   & 1  & 3.98E-19  & TRUE  \\\hline
Llama-8B & -1   & 0.267  & 236.164   & 1  & 1.87E-27  & TRUE  \\\hline
Llama-8B & 0    & 1.478  & 2826.097  & 1  & 1.72E-73  & TRUE  \\\hline
Llama-8B & 1    & 0.666  & 985.609   & 1  & 1.88E-52  & TRUE  \\\hline
Llama-8B & 2    & 0.573  & 315.491   & 1  & 6.53E-32  & TRUE  \\\hline
Llama-8B & 3    & 0.418  & 164.376   & 1  & 2.42E-22  & TRUE  \\\hline
Llama-8B & 4    & 0.312  & 79.037    & 1  & 5.33E-14  & TRUE  \\\hline
Llama-8B & 5    & 0.162  & 19.22     & 1  & 3.78E-05  & TRUE  \\\hline
Llama-8B & 6    & 0.151  & 19.336    & 1  & 3.78E-05  & TRUE 
\\ \hline\end{tabular}
\end{table}

\begin{table}[H]
\caption{Statistics to accompany Figure 5D. Linear models testing the association between number of items until a cluster and activation of anticipatory loading}
\label{tab:ramping}
\small
\begin{tabular}{|l|l|l|l|l|}\hline
model    & beta   & chi2    & p         & sig  \\\hline
Gemma-9B & -0.316 & 525.373 & 1.43E-115 & TRUE \\\hline
Gemma-2B & -0.239 & 179.959 & 1.24E-40  & TRUE \\\hline
Qwen-7B  & -0.105 & 30.040  & 5.29E-08  & TRUE \\\hline
Llama-3B & -0.079 & 20.439  & 6.16E-06  & TRUE \\\hline
Llama-8B & -0.104 & 67.895  & 2.87E-16  & TRUE
\\ \hline\end{tabular}
\end{table}

{\small
\begin{longtable}{|l|l|l|l|l|l|}
\caption{Statistics to accompany Figure 5E. Fisher tests comparing switch rates during positive steering against noise.}
\label{tab:positive_steering}\\
\hline
model    & layer & base switch rate& steer switch rate& noise switch rate& p         \\\hline
Gemma-9B & 2     & 0.29                   & 0.27                & 0.27                & 1.00E+00  \\\hline
Gemma-9B & 4     & 0.29                   & 0.32                & 0.26                & 1.15E-12  \\\hline
Gemma-9B & 6     & 0.29                   & 0.32                & 0.27                & 1.93E-11  \\\hline
Gemma-9B & 8     & 0.29                   & 0.32                & 0.28                & 4.34E-09  \\\hline
Gemma-9B & 10    & 0.29                   & 0.36                & 0.30                & 5.06E-11  \\\hline
Gemma-9B & 12    & 0.29                   & 0.45                & 0.34                & 1.15E-31  \\\hline
Gemma-9B & 14    & 0.29                   & 0.51                & 0.36                & 4.67E-50  \\\hline
Gemma-9B & 16    & 0.29                   & 0.58                & 0.42                & 6.26E-51  \\\hline
Gemma-9B & 18    & 0.29                   & 0.58                & 0.34                & 2.25E-121 \\\hline
Gemma-9B & 20    & 0.29                   & 0.75                & 0.31                & 3.84E-292 \\\hline
Gemma-9B & 22    & 0.29                   & 0.79                & 0.35                & 5.00E-301 \\\hline
Gemma-9B & 24    & 0.29                   & 0.81                & 0.37                & 5.43E-241 \\\hline
Gemma-9B & 26    & 0.29                   & 0.70                & 0.29                & 0.00E+00  \\\hline
Gemma-9B & 28    & 0.29                   & 0.64                & 0.28                & 5.32E-298 \\\hline
Gemma-9B & 30    & 0.29                   & 0.47                & 0.28                & 9.18E-110 \\\hline
Gemma-9B & 32    & 0.29                   & 0.38                & 0.26                & 2.88E-63  \\\hline
Gemma-9B & 34    & 0.29                   & 0.35                & 0.22                & 1.24E-70  \\\hline
Gemma-9B & 36    & 0.29                   & 0.32                & 0.22                & 1.95E-37  \\\hline
Gemma-9B & 38    & 0.29                   & 0.34                & 0.26                & 5.91E-22  \\\hline
Gemma-9B & 40    & 0.29                   & 0.32                & 0.29                & 1.98E-04  \\\hline
Gemma-2B & 2     & 0.35                   & 0.57                & 0.54                & 1.66E-01  \\\hline
Gemma-2B & 4     & 0.35                   & 0.61                & 0.41                & 4.94E-42  \\\hline
Gemma-2B & 6     & 0.35                   & 0.66                & 0.62                & 8.63E-02  \\\hline
Gemma-2B & 8     & 0.35                   & 0.69                & 0.48                & 1.26E-41  \\\hline
Gemma-2B & 10    & 0.35                   & 0.72                & 0.43                & 5.91E-111 \\\hline
Gemma-2B & 12    & 0.35                   & 0.81                & 0.42                & 8.24E-142 \\\hline
Gemma-2B & 14    & 0.35                   & 0.83                & 0.23                & 1.07E-190 \\\hline
Gemma-2B & 16    & 0.35                   & 0.76                & 0.31                & 4.77E-209 \\\hline
Gemma-2B & 18    & 0.35                   & 0.47                & 0.21                & 4.92E-182 \\\hline
Gemma-2B & 20    & 0.35                   & 0.49                & 0.30                & 1.05E-63  \\\hline
Gemma-2B & 22    & 0.35                   & 0.40                & 0.34                & 1.77E-08  \\\hline
Gemma-2B & 24    & 0.35                   & 0.59                & 0.39                & 6.22E-57  \\\hline
Qwen-7B  & 2     & 0.40                   & 0.32                & 0.33                & 1.00E+00  \\\hline
Qwen-7B  & 4     & 0.40                   & 0.40                & 0.21                & 1.05E-51  \\\hline
Qwen-7B  & 6     & 0.40                   & 0.24                & 0.22                & 1.45E-01  \\\hline
Qwen-7B  & 8     & 0.40                   & 0.71                & 0.48                & 3.10E-46  \\\hline
Qwen-7B  & 10    & 0.40                   & 0.78                & 0.57                & 1.07E-24  \\\hline
Qwen-7B  & 12    & 0.40                   & 0.82                & 0.59                & 4.34E-43  \\\hline
Qwen-7B  & 14    & 0.40                   & 0.81                & 0.27                & 2.11E-241 \\\hline
Qwen-7B  & 16    & 0.40                   & 0.82                & 0.34                & 1.78E-193 \\\hline
Qwen-7B  & 18    & 0.40                   & 0.85                & 0.33                & 8.99E-244 \\\hline
Qwen-7B  & 20    & 0.40                   & 0.83                & 0.62                & 4.44E-44  \\\hline
Qwen-7B  & 22    & 0.40                   & 0.81                & 0.55                & 5.98E-61  \\\hline
Qwen-7B  & 24    & 0.40                   & 0.66                & 0.58                & 1.10E-08  \\\hline
Qwen-7B  & 26    & 0.40                   & 0.44                & 0.51                & 1.00E+00  \\\hline
Llama-8B & 2     & 0.32                   & 0.33                & 0.44                & 1.00E+00  \\\hline
Llama-8B & 4     & 0.32                   & 0.58                & 0.35                & 1.92E-88  \\\hline
Llama-8B & 6     & 0.32                   & 0.66                & 0.37                & 1.50E-109 \\\hline
Llama-8B & 8     & 0.32                   & 0.52                & 0.48                & 3.22E-03  \\\hline
Llama-8B & 10    & 0.32                   & 0.76                & 0.61                & 2.07E-47  \\\hline
Llama-8B & 12    & 0.32                   & 0.70                & 0.62                & 1.41E-06  \\\hline
Llama-8B & 14    & 0.32                   & 0.63                & 0.60                & 2.06E-01  \\\hline
Llama-8B & 16    & 0.32                   & 0.44                & 0.44                & 8.33E-01  \\\hline
Llama-8B & 18    & 0.32                   & 0.42                & 0.45                & 1.00E+00  \\\hline
Llama-8B & 20    & 0.32                   & 0.42                & 0.56                & 1.00E+00  \\\hline
Llama-8B & 22    & 0.32                   & 0.44                & 0.44                & 1.00E+00  \\\hline
Llama-8B & 24    & 0.32                   & 0.61                & 0.48                & 1.47E-24  \\\hline
Llama-8B & 26    & 0.32                   & 0.61                & 0.43                & 1.68E-57  \\\hline
Llama-8B & 28    & 0.32                   & 0.56                & 0.44                & 2.53E-26  \\\hline
Llama-8B & 30    & 0.32                   & 0.55                & 0.47                & 8.83E-17  \\\hline
Llama-3B & 2     & 0.54                   & 0.63                & 0.53                & 8.86E-01  \\\hline
Llama-3B & 4     & 0.54                   & 0.66                & 0.44                & 3.66E-12  \\\hline
Llama-3B & 6     & 0.54                   & 0.71                & 0.61                & 2.74E-03  \\\hline
Llama-3B & 8     & 0.54                   & 0.66                & 0.57                & 2.68E-03  \\\hline
Llama-3B & 10    & 0.54                   & 0.80                & 0.75                & 2.56E-01  \\\hline
Llama-3B & 12    & 0.54                   & 0.74                & 0.71                & 5.83E-01  \\\hline
Llama-3B & 14    & 0.54                   & 0.59                & 0.73                & 1.00E+00  \\\hline
Llama-3B & 16    & 0.54                   & 0.66                & 0.41                & 6.11E-21  \\\hline
Llama-3B & 18    & 0.54                   & 0.58                & 0.59                & 1.00E+00  \\\hline
Llama-3B & 20    & 0.54                   & 0.56                & 0.44                & 4.83E-10  \\\hline
Llama-3B & 22    & 0.54                   & 0.53                & 0.53                & 8.71E-01  \\\hline
Llama-3B & 24    & 0.54                   & 0.61                & 0.49                & 3.68E-10  \\\hline
Llama-3B & 26    & 0.54                   & 0.53                & 0.52                & 4.43E-01 
\end{longtable}
}
\newpage

{\small
\begin{longtable}{|l|l|l|l|l|l|}
\caption{Statistics to accompany Figure 5F. Fisher tests comparing switch rates during negative steering against noise}
\label{tab:negative_steering_stats}\\
\hline
model    & layer & base switch rate & steer switch rate & noise switch rate & p         \\\hline
Gemma-9B & 2     & 0.29             & 0.26              & 0.28              & 3.30E-02  \\\hline
Gemma-9B & 4     & 0.29             & 0.25              & 0.28              & 1.51E-05  \\\hline
Gemma-9B & 6     & 0.29             & 0.25              & 0.29              & 1.70E-06  \\\hline
Gemma-9B & 8     & 0.29             & 0.26              & 0.28              & 1.15E-01  \\\hline
Gemma-9B & 10    & 0.29             & 0.28              & 0.32              & 4.17E-05  \\\hline
Gemma-9B & 12    & 0.29             & 0.32              & 0.32              & 7.75E-01  \\\hline
Gemma-9B & 14    & 0.29             & 0.36              & 0.31              & 1.00E+00  \\\hline
Gemma-9B & 16    & 0.29             & 0.45              & 0.37              & 1.00E+00  \\\hline
Gemma-9B & 18    & 0.29             & 0.53              & 0.29              & 1.00E+00  \\\hline
Gemma-9B & 20    & 0.29             & 0.62              & 0.46              & 1.00E+00  \\\hline
Gemma-9B & 22    & 0.29             & 0.46              & 0.45              & 1.00E+00  \\\hline
Gemma-9B & 24    & 0.29             & 0.25              & 0.40              & 1.26E-40  \\\hline
Gemma-9B & 26    & 0.29             & 0.07              & 0.32              & 1.17E-274 \\\hline
Gemma-9B & 28    & 0.29             & 0.18              & 0.42              & 2.20E-160 \\\hline
Gemma-9B & 30    & 0.29             & 0.23              & 0.46              & 1.62E-166 \\\hline
Gemma-9B & 32    & 0.29             & 0.26              & 0.36              & 3.48E-42  \\\hline
Gemma-9B & 34    & 0.29             & 0.17              & 0.31              & 1.23E-102 \\\hline
Gemma-9B & 36    & 0.29             & 0.15              & 0.25              & 2.86E-78  \\\hline
Gemma-9B & 38    & 0.29             & 0.21              & 0.29              & 2.48E-30  \\\hline
Gemma-9B & 40    & 0.29             & 0.29              & 0.29              & 1.00E+00  \\\hline
Gemma-2B & 2     & 0.35             & 0.39              & 0.26              & 1.00E+00  \\\hline
Gemma-2B & 4     & 0.35             & 0.30              & 0.29              & 1.00E+00  \\\hline
Gemma-2B & 6     & 0.35             & 0.45              & 0.54              & 5.21E-16  \\\hline
Gemma-2B & 8     & 0.35             & 0.16              & 0.41              & 7.74E-153 \\\hline
Gemma-2B & 10    & 0.35             & 0.25              & 0.47              & 1.94E-87  \\\hline
Gemma-2B & 12    & 0.35             & 0.06              & 0.61              & 0.00E+00  \\\hline
Gemma-2B & 14    & 0.35             & 0.00              & 0.64              & 0.00E+00  \\\hline
Gemma-2B & 16    & 0.35             & 0.15              & 0.61              & 8.03E-264 \\\hline
Gemma-2B & 18    & 0.35             & 0.25              & 0.53              & 3.80E-95  \\\hline
Gemma-2B & 20    & 0.35             & 0.44              & 0.47              & 8.07E-02  \\\hline
Gemma-2B & 22    & 0.35             & 0.34              & 0.43              & 2.86E-10  \\\hline
Gemma-2B & 24    & 0.35             & 0.25              & 0.42              & 2.18E-57  \\\hline
Qwen-7B  & 2     & 0.40             & 0.47              & 0.48              & 4.23E-01  \\\hline
Qwen-7B  & 4     & 0.40             & 0.39              & 0.42              & 3.92E-02  \\\hline
Qwen-7B  & 6     & 0.40             & 0.49              & 0.56              & 4.50E-07  \\\hline
Qwen-7B  & 8     & 0.40             & 0.51              & 0.56              & 5.54E-04  \\\hline
Qwen-7B  & 10    & 0.40             & 0.44              & 0.66              & 1.28E-52  \\\hline
Qwen-7B  & 12    & 0.40             & 0.26              & 0.36              & 9.82E-19  \\\hline
Qwen-7B  & 14    & 0.40             & 0.35              & 0.51              & 4.78E-33  \\\hline
Qwen-7B  & 16    & 0.40             & 0.21              & 0.60              & 4.50E-200 \\\hline
Qwen-7B  & 18    & 0.40             & 0.16              & 0.47              & 7.17E-173 \\\hline
Qwen-7B  & 20    & 0.40             & 0.21              & 0.25              & 2.39E-06  \\\hline
Qwen-7B  & 22    & 0.40             & 0.25              & 0.33              & 1.79E-13  \\\hline
Qwen-7B  & 24    & 0.40             & 0.27              & 0.34              & 1.25E-08  \\\hline
Qwen-7B  & 26    & 0.40             & 0.35              & 0.42              & 5.56E-09  \\\hline
Llama-8B & 2     & 0.32             & 0.80              & 0.35              & 1.00E+00  \\\hline
Llama-8B & 4     & 0.32             & 0.45              & 0.65              & 4.79E-19  \\\hline
Llama-8B & 6     & 0.32             & 0.43              & 0.67              & 1.07E-66  \\\hline
Llama-8B & 8     & 0.32             & 0.53              & 0.63              & 2.30E-14  \\\hline
Llama-8B & 10    & 0.32             & 0.51              & 0.56              & 6.82E-03  \\\hline
Llama-8B & 12    & 0.32             & 0.48              & 0.55              & 8.91E-06  \\\hline
Llama-8B & 14    & 0.32             & 0.58              & 0.31              & 1.00E+00  \\\hline
Llama-8B & 16    & 0.32             & 0.44              & 0.43              & 1.00E+00  \\\hline
Llama-8B & 18    & 0.32             & 0.52              & 0.56              & 3.08E-03  \\\hline
Llama-8B & 20    & 0.32             & 0.44              & 0.50              & 1.66E-04  \\\hline
Llama-8B & 22    & 0.32             & 0.32              & 0.44              & 7.64E-21  \\\hline
Llama-8B & 24    & 0.32             & 0.39              & 0.47              & 1.82E-10  \\\hline
Llama-8B & 26    & 0.32             & 0.39              & 0.44              & 2.05E-05  \\\hline
Llama-8B & 28    & 0.32             & 0.29              & 0.36              & 7.93E-12  \\\hline
Llama-8B & 30    & 0.32             & 0.28              & 0.36              & 4.38E-15  \\\hline
Llama-3B & 2     & 0.54             &                   & 0.57              & NA        \\\hline
Llama-3B & 4     & 0.54             & 0.65              & 0.46              & 1.00E+00  \\\hline
Llama-3B & 6     & 0.54             & 0.53              & 0.57              & 8.12E-02  \\\hline
Llama-3B & 8     & 0.54             & 0.75              & 0.60              & 1.00E+00  \\\hline
Llama-3B & 10    & 0.54             & 0.29              & 0.60              & 2.04E-36  \\\hline
Llama-3B & 12    & 0.54             & 0.26              & 0.45              & 3.10E-47  \\\hline
Llama-3B & 14    & 0.54             & 0.33              & 0.65              & 1.16E-124 \\\hline
Llama-3B & 16    & 0.54             & 0.48              & 0.65              & 4.83E-33  \\\hline
Llama-3B & 18    & 0.54             & 0.44              & 0.62              & 6.81E-45  \\\hline
Llama-3B & 20    & 0.54             & 0.63              & 0.61              & 1.00E+00  \\\hline
Llama-3B & 22    & 0.54             & 0.54              & 0.52              & 1.00E+00  \\\hline
Llama-3B & 24    & 0.54             & 0.41              & 0.51              & 4.52E-13  \\\hline
Llama-3B & 26    & 0.54             & 0.34              & 0.57              & 1.82E-36 
\end{longtable}
}
\newpage

\end{document}